\documentclass[sn-mathphys,Numbered]{sn-jnl}% Math and Physical Sciences Reference Style
\usepackage{graphicx}%
\usepackage{multirow}%
\usepackage{amsmath,amssymb,amsfonts}%
\usepackage{amsthm}%
\usepackage{mathrsfs}%
\usepackage[title]{appendix}%
\usepackage{xcolor}%
\usepackage{textcomp}%
\usepackage{manyfoot}%
\usepackage{booktabs}%
\usepackage{algorithm}%
\usepackage{algorithmicx}%
\usepackage{algpseudocode}%
\usepackage{listings}%

\usepackage{parskip}
\usepackage{dsfont}
\usepackage{mathtools}%
\usepackage{subcaption}%
\usepackage{adjustbox}%
\usepackage[boldmath]{numprint}

\newcommand{\mycomment}[1]{}
\usepackage{float}
\floatstyle{plaintop}
\restylefloat{table}
\usepackage{longtable}
\usepackage{booktabs}
\usepackage{makecell}
\usepackage{orcidlink}
\theoremstyle{thmstyleone}%
\theoremstyle{thmstyletwo}%

\theoremstyle{thmstylethree}%

\begin{document}

\title[Article Title]{Multi-SpaCE: Multi-Objective Subsequence-based Sparse Counterfactual Explanations for Multivariate Time Series Classification}


\author*[1]{\fnm{Mario} \sur{Refoyo} \orcidlink{0009-0001-4087-930X}}\email{m.refoyo@upm.es}

\author[1]{\fnm{David} \sur{Luengo} \orcidlink{0000-0001-7407-3630}}\email{david.luengo@upm.es}

\affil[1]{\orgdiv{Department of Audiovisual and Communications Engineering}, \orgname{Universidad Politécnica de Madrid (UPM)}, \orgaddress{\street{Calle Nikola Tesla s/n}, \city{Madrid}, \postcode{28031}, \country{Spain}}}


\abstract{Deep Learning systems excel in complex tasks but often lack transparency, limiting their use in critical applications. Counterfactual explanations, a core tool within eXplainable Artificial Intelligence (XAI), offer insights into model decisions by identifying minimal changes to an input to alter its predicted outcome. However, existing methods for time series data are limited by univariate assumptions, rigid constraints on modifications, or lack of validity guarantees. This paper introduces Multi-SpaCE, a multi-objective counterfactual explanation method for multivariate time series. Using non-dominated ranking genetic algorithm II (NSGA-II), Multi-SpaCE balances proximity, sparsity, plausibility, and contiguity. Unlike most methods, it ensures perfect validity, supports multivariate data and provides a Pareto front of solutions, enabling flexibility to different end-user needs. Comprehensive experiments in diverse datasets demonstrate the ability of Multi-SpaCE to consistently achieve perfect validity and deliver superior performance compared to existing methods.}

\keywords{eXplainable Artificial Intelligence (XAI), Counterfactual Explanations, Genetic Algorithm Optimization, Time Series Classification}



\maketitle

\newpage
\section{Introduction}
\label{sec:intro}

Machine Learning in general, and particularly Deep Learning, is becoming increasingly pervasive, automating tasks and improving decision-making across industries such as healthcare, finance, and manufacturing \cite{sarker2022applicationsai}.
However, despite their benefits, Deep Learning systems often face critical challenges, such as their ``black-box'' nature.
These systems excel in complex tasks but lack transparency, making it difficult for humans to interpret their decisions.
This opacity hinders their deployment in high-stakes applications, where understanding the rationale behind decisions is crucial.
To address this issue, eXplainable Artificial Intelligence (XAI) has emerged as a field dedicated to making machine learning models more interpretable, fair, and transparent \cite{barredo2020survey,molnar2022}.
By enabling decision-makers to trust AI systems, XAI aims to facilitate the integration of AI into critical domains. 

Among the many explanation techniques in XAI, counterfactual explanations (CFEs) \cite{wachter2017} have gained significant attention due to their alignment with human cognitive processes \cite{byrne2019, MILLER2019}.
CFEs identify the smallest changes required by an input instance to achieve a desired outcome, thus allowing users to explore ``what-if'' scenarios.
Originally framed as optimization problems, CFEs searched for proximity between the original instance and the counterfactual, while ensuring the validity of the counterfactual (the change in the outcome) \cite{guidotti2022, verma2024}.
With time, counterfactual explanations have evolved to incorporate desiderata such as plausibility \cite{dhurandhar2018, vanlooveren2019, lang2022sparse}, diversity \cite{mothilal2019}, and actionability \cite{karimi2020recourse}.
As new desiderata were identified, the conflicts between them also became apparent.
Multi-objective optimization has become an effective tool to address these trade-offs between conflicting properties \cite{dandl2020moc, barredo2020moc, hollig2022tsevo}, providing users with a Pareto front of solutions that can be easily adapted to their preferences.

While CFEs were initially developed for tabular data, their adaptation to time series remains an active area of research.
Existing methods typically leverage subsequences of changes to address the higher dimensionality and serial correlations inherent in time series inputs
\cite{delaney2021, ates2021, guidotti2020lasts, spinnato2023lastsv2, wang2024glacier, li2023abcf,  bahri2024discord, hollig2022tsevo, huang2024txgen, refoyo2024subspace}.
However, most methods also enforce rigid constraints, such as single or fixed-length subsequences, limiting their flexibility and effectiveness.
Furthermore, many approaches remain focused on univariate settings, failing to account for the complexities of the multivariate time series which are commonly found in real-world applications.
Most importantly, validity is frequently treated as another objective rather than a strict requirement.
While many works define validity as a constraint, either verbally or mathematically \cite{guidotti2022}, the heuristic approaches used to solve these problems often fail to enforce it.
We argue that validity is fundamental in CFEs and should be treated as a strict requirement.
Without validity, all the other performance metrics lose their significance, as they are no longer associated with explanations that achieve the desired outcome.
Methods that fail to enforce validity might be unsuitable for practical applications.

With this analysis, we argue that i) Multi-objective obtimization is desired, as it offers a natural framework to address conflicting properties; ii) methods should support multivariate time series data to ensure applicability to a wide range of real-world problems; and iii) the validity of counterfactual explanations should be treated as a strict requirement rather than just another metric to optimize. 

\begin{figure}[h]
    \centering
    \includegraphics[width=1\textwidth]{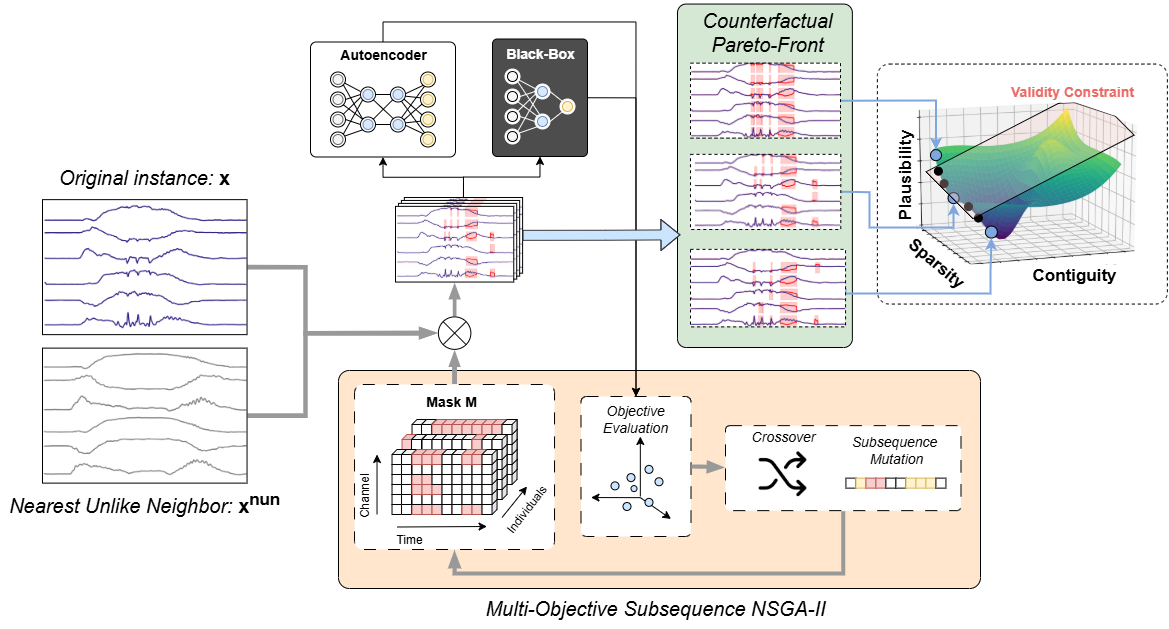}
    \caption{Multi-SpaCE block diagram, including the black-box classifier to be explained, and the autoencoder used to measure the plausibility of the generated CFEs.}
    \label{fig:block_diagram}
\end{figure}

These three principles form the foundation of our proposed method, Multi-SpaCE, which addresses these challenges by ensuring the validity of the returned counterfactuals, supporting multivariate time series, and leveraging a multi-objective optimization framework.
Multi-SpaCE extends our previous work, Sub-SpaCE \cite{refoyo2024subspace}, which uses a genetic algorithm (GA) with custom mutation and initialization to generate counterfactual explanations by optimizing sparsity, plausibility and minimizing the number of subsequences.
Multi-SpaCE inherits all the good properties of Sub-SpaCE (such as its model-agnostic nature, its computational efficiency or the validity of all the generated CFEs), while overcoming its two major limitations: its restriction to univariate data and its reliance on extensive experimentation to balance multiple desiderata.
By introducing a multi-objective optimization framework based on Non-Dominated Sorting Genetic Algorithm II (NSGA-II) \cite{deb2002nsga}, Multi-SpaCE offers a Pareto-Front of solutions, enabling users to explore trade-offs between objectives.
The optimization task identifies the points in the original input to modify, substituting their values with those of the Nearest Unlike Neighbor (NUN) \cite{delaney2021}.
%
%Critically, Multi-SpaCE guarantees the generation of valid counterfactuals, unlike most existing methods.
%
Figure~\ref{fig:block_diagram} presents the architecture of the proposed solution: the NUNs obtained for each input instance, the black box classifier to be explained, the NSGA algorithm used to generate candidate CFEs, the Autoencoder introduced to measure their plausibility, and the Pareto-front of counterfactuals returned by Multi-SpaCE.

\vspace*{-6pt}

To the best of our knowledge, Multi-SpaCE is the first multi-objective optimization method that ensures perfectly valid counterfactual explanations for multivariate time series classification problems.
Extensive experimentation demonstrates its ability to achieve perfect validity across different datasets while outperforming existing methods in multiple metrics.

\section{Related work}
\label{sec:related_work}

The field of eXplainable Artificial Intelligence (XAI) has experienced significant growth in recent years, resulting in the development of numerous methods tailored to specific domains and data types \cite{saeed2023metasurvey}.
However, XAI methods for time series have received considerably less attention than those designed for tabular or image data \cite{rojat2021TSsurvey, vielhaben2024TSInspectionLayers}.
Many existing methods are adaptations of those designed for tabular or image data \cite{theissler2022explainable}, and cannot properly cope with the unique characteristics of time-ordered inputs \cite{ismail2020}.
Time series data, such as electroencephalograms (EEG), electrocardiograms (ECG), human motion, or financial signals, require domain expertise to interpret, making traditional methods less effective to generate meaningful explanations \cite{rojat2021TSsurvey, olivas2024HM}.

This tendency extends to counterfactual explanations. Formally, a counterfactual explanation $x'$ is the smallest variation of an original instance $x$ that changes the predicted outcome $y=b(x)$ of a black-box classifier $b\!: X \rightarrow [0,1]$, subject to $b(x') \neq y$ \cite{guidotti2022}.
Most counterfactual explanation methods are designed for tabular data. When applied to time series, these methods may not scale well with the increased dimensionality of inputs and often fail to account for the serial correlations that characterize time series data \cite{delaney2021}. 

To address these challenges, methods such as Native Guide (NG) \cite{delaney2021} introduced contiguity, ensuring that changes occur as subsequences aligned with the sequential structure of time series.
NG finds the Nearest Unlike Neighbor (NUN), the closest example to the instance being explained that belongs to a different class, and substitutes the shortest subsequence of the original instance with the corresponding subsequence from the NUN to achieve the desired outcome.
The main drawback of NG is that it is limited to univariate settings and allows only a single subsequence of changes, thus restricting its practical applicability.
COMTE, by Ates et al. \cite{ates2021}, extended this idea to multivariate time series by optimizing the number of channels substituted from the NUN to produce a valid counterfactual.
However, this approach is still limited to a single subsequence per channel, which may introduce unnecessary changes and break the existing relationships between channels. 

Other methods leverage shapelets to guide counterfactual explanations.
For example, Local Agnostic Subsequence-based Time Series Explainer (LASTS) \cite{guidotti2020lasts, spinnato2023lastsv2} uses the latent space of a Variational Autoencoder (VAE) to generate a set of exemplars and counterfactuals in the neighborhood of the instance to explain.
It then derives classification rules based on the presence of specific shapelets.
Another work, Shapelet Explainer for Time Series (SETS) \cite{bahri2022-SETS}, extracts discriminative shapelets for each class.
It then localizes them in the input to explain, and substitutes their corresponding values with those of the NUN. %While effective, However, these methods rely on predetermined shapelet lengths, and not optimize the number of them used.

More recent approaches, such as Attention-Based Counterfactual Explanations (AB-CF) \cite{li2023abcf} and Discord-based Counterfactual Explanations (DiscoX) \cite{bahri2024discord}, address multivariate time series but rely on fixed-length subsequences.
AB-CF greedily selects fixed-length multichannel subsequences that maximize classification entropy when isolated, while DiscoX identifies fixed-length discords (subsequences maximally distant from their nearest neighbors in the desired class) and replaces them with the nearest subsequences from the desired output class.
Although these methods perform well in some scenarios, their reliance on fixed-length subsequences often results in unnecessary changes.
Additionally, their heuristic and greedy approaches reduce their validity rates, limiting their applicability in real-world scenarios.

Another line of research still follows the optimization-based counterfactual generation originally proposed by Wachter et al. \cite{wachter2017}.
Building on this foundational work, Glacier \cite{wang2024glacier} generates counterfactuals in the latent space of an autoencoder while penalizing changes outside relevant subsequences identified by LIMESegment \cite{sivill2022limesegment}.
However, Glacier treats validity as a loss term rather than a strict constraint, leading to low validity rates, and remains limited to univariate datasets.
Genetic algorithms (GAs) have also been applied to generate CFEs.
TSEvo \cite{hollig2022tsevo} introduced a multi-objective approach and diverse mutation operators to optimize both the location and magnitude of changes.
However, TSEvo’s high-dimensional search space results in long execution times and low validity rates.
More recently, TX-Gen \cite{huang2024txgen} has tackled multi-objective optimization by finding a single subsequence of changes.
Instead of relying on the NUN, TX-Gen employs a compressed representation of the subsequence and uses a simple AR model to generate the content of the subsequences proposed by the optimizer.
Although TX-Gen reports perfect validity, it was only tested in univariate settings and might struggle with plausibility in more complex scenarios, where a simple generative AR model might not provide enough flexibility.
Finally, Sub-SpaCE \cite{refoyo2024subspace} balances sparsity, plausibility, and the number of subsequences by framing counterfactual generation as a single-objective optimization problem solved using a GA with custom mutation and initialization.
Sub-SpaCE identifies a binary mask of changes, indicating the locations where the values of the original instance should be substituted with those of the NUN.
While effective, Sub-SpaCE is restricted to univariate data and lacks a multi-objective framework, requiring experimentation to balance objective relevance.

As stated above, contiguity is a desired property, as grouped changes in the form of subsequences enhance interpretability.
However, current methods either promote contiguity indirectly (e.g., TSEvo by using the mutation operators), or limit their flexibility by imposing: i) a single subsequence of changes (e.g., NG, COMTE and TX-Gen) or; ii) fixed-length subsequences (e.g., AB-CF and DiscoX), thus hindering their adaptability to complex scenarios.
Additionally, methods should be prepared to work in multivariate settings, which are ubiquitous in real-world applications, something that NG, Glacier, TX-Gen, and Sub-SpaCE do not support.
Finally, and most importantly, except for a few approaches (such as NG, TX-Gen and Sub-SpaCE), most methods fail to ensure perfect validity, treating it as another objective rather than as a strict constraint.

\section{Methods}
\label{sec:methodology}

\subsection{Sub-SpaCE}

Sub-SpaCE is a recent method developed to generate sparse and plausible counterfactual explanations for univariate time series classification problems by changing only a few relevant subsequences of the original instance \cite{refoyo2024subspace}.
It formulates the generation of counterfactual explanations as an optimization problem, which is solved by using a Genetic Algorithm (GA) that balances all the terms that are considered relevant: sparsity, plausibility, contiguity, and validity.
It uses customized initialization and mutation processes to improve the convergence properties of a vanilla genetic algorithm. 

Let $\mathbf{x} = \{x_1, x_2, \dots, x_L\} \in \mathbb{R}^L$ represent the univariate time series to explain, where $L$ is the length of the series, and let $y=b(x)$ be the output class of the black-box model $b$ for that instance.
Sub-SpaCE generates a counterfactual $\mathbf{x}'$ by applying a mask of changes $\mathbf{m} \in \{0,1\}^L$, which specifies the indices at which the original values of $x$ are replaced by the corresponding values from the ``Nearest Unlike Neighbor'' (NUN), $\mathbf{x}^{nun}$.
The NUN is defined as the instance from a different class, $y^{nun} \ne y$, that is the closest to $\mathbf{x}$ in Euclidean distance \cite{delaney2021}.
The counterfactual is then
\[
x'_i = f(x_i|m_i, x_i^{nun}) =  
\begin{cases}
   x^{nun}_i,  &\quad\text{if } m_i = 1;\\
   x_i,  &\quad\text{else.} \\
\end{cases}
\quad \forall i \in \{1, \dots, L\}.
\]
The objective is to find the $\mathbf{m}$ that minimizes the number of changes while achieving the desired NUN class, $y^{nun}$.
The optimization is composed of several normalized loss terms, each one aiming to achieve a desiderata of counterfactual explanations:
\begin{itemize}
    \item \textbf{Adversarial Loss}: $L_{adv} = p_b(\mathbf{x'}, y^{nun})$, where $p_b(\mathbf{x'}, y^{nun})$ is the classifier’s probability for the desired class $y^{nun}$ given $\mathbf{x'}$. This term increases the likelihood of the counterfactual belonging to the desired class. It helps the search to achieve valid counterfactual solutions.
    \item \textbf{Sparsity Loss}: $L_{spa} = \frac{||\mathbf{m}||_0}{L}$, where $||\mathbf{m}||_0$ is the $\ell_0$ pseudo norm. To minimize changes, Sub-SpaCE penalizes the number of non-zero elements in $\mathbf{m}$ (i.e., positions where $m_i = 1$ and the original value is substituted with the value of the NUN).
    \item \textbf{Contiguity Loss}: Measured as the number of subsequences $L_{sub} = \left( \frac{\sum_{i=2}^{L} \mathds{1}_{i}}{L/2} \right)^{\gamma}$, where $\mathds{1}_{i} = 1$ when a new subsequence begins (i.e., $m_{i-1} = 0$ and $m_i = 1$). This term minimizes the number of contiguous subsequences, promoting interpretability and reducing the complexity of the explanation. The hyperparameter $\gamma$ controls the penalty's convexity, encouraging fewer subsequences when $\gamma < 1$. During the experiments, the value of $\gamma$ was set to $0.25$ to encourage a higher penalization for a lower number of subsequences.
    \item \textbf{Plausibility Loss}: measured as the Increase in Outlier Score Loss $L_{ios} = \frac{e(\mathbf{x}, \mathbf{x'})}{e_{\max}}$, where $e(\mathbf{x}, \mathbf{x'}) = \min(0, ||\mathbf{x'} - f_{AE}(\mathbf{x'})||_2 - ||\mathbf{x} - f_{AE}(\mathbf{x})||_2)$ and $e_{\max}$ is the maximum reconstruction error on the training set. This term enforces plausibility by penalizing deviations from the original data manifold. Using an autoencoder $f_{AE}$, trained to reconstruct the data, this term measures how far $\mathbf{x'}$ is from the data distribution compared to the original instance $\mathbf{x}$. This metric guides the search to counterfactuals as plausible as the original instance.
\end{itemize}

% To solve the optimization problem, Sub-SpaCE resorts to a Genetic Algorithm (GA) that adapts the objective function in \eqref{eq:cf_complete_loss} to the fitness function to be maximized: 
To solve the optimization problem, Sub-SpaCE resorts to a Genetic Algorithm (GA) that tries to maximize the following fitness function: 
\begin{equation}\label{eq:cf_ts_fitness}
\mathcal{F}(\mathbf{m}) = \alpha L_{adv} - \beta L_{spa} - \eta L_{sub} - \lambda L_{ios} - \nu \cdot \mathds{1}_{class}(\mathbf{x'}, \mathbf{m}, y^{nun}),
\end{equation}
where $\mathds{1}_{class}$ is an indicator function penalizing classifications that differ from the desired class $y^{nun}$, and $\nu$ is a large scalar to strictly enforce this constraint.
Each term is scaled to a range of $[0, 1]$, making it easier to balance with the parameters $\alpha$, $\beta$, $\eta$, and $\lambda$, which sum to one.
The GA starts with an initial population $\mathcal{P} = \{\mathbf{m}_0, \mathbf{m}_1, \dots \mathbf{m}_N\}$ of $N$ candidate masks, and iteratively updates it to maximize  $\mathcal{F}(m)$ through parent selection, crossover, and mutation.
To improve the results, Sub-SpaCE proposes two modifications that improve both results and convergence properties of the algorithm:
\begin{itemize}
    \item \textbf{Initialization:} A feature-attribution method identifies crucial parts of $x$ to prioritize during initialization. This importance is combined with Gaussian noise to diversify the population, thus allowing the algorithm to start from better than random solutions. Then it sets to 1 the $h\%$ most activated values and sets to 0 the rest. Sub-SpaCE also implements a reinitialization process that restarts the optimization with an increased number of activated values, $h_{inc}$, if a valid solution was not found within the first $G_{reinit}$ iterations.
    \item \textbf{Subsequence-based Mutation:} To smooth the fitness scores during the iterative process, mutations only extend or shorten existing subsequences in the population, instead of randomly changing the values of $\mathbf{m}$. This proved to enhance stability while speeding convergence.
\end{itemize}

\subsection{Multi-SpaCE}

Sub-SpaCE is a method designed for generating counterfactual explanations for univariate time series classification problems. While it achieves great results in its target domain, Sub-SpaCE has two primary limitations that hinder its broader applicability:

\begin{itemize}
    \item \textbf{Limited to univariate time series:} Sub-SpaCE does not support multivariate time series, which are prevalent in real-world scenarios, such as healthcare, finance, or sensor networks. Multivariate datasets introduce additional complexity, since changes across multiple channels (variables) must be coordinated.
    \item \textbf{Utility function tuning:} Sub-SpaCE relies on a utility function with multiple hyperparameters $(\alpha, \beta, \eta, \lambda)$ to balance its loss terms. Although these terms are normalized to $[0, 1]$, finding the appropriate balance often requires trial-and-error experimentation, which can be time-consuming and user-dependent.
\end{itemize}

To overcome these limitations, Multi-SpaCE extends Sub-SpaCE to support multivariate time series, including new mutation operators and reframing the optimization as a multi-objective problem.
Multi-SpaCE provides a Pareto front of non-dominated solutions, allowing users to explore trade-offs between objectives and select explanations aligned with their specific needs without the need for experimentation.
Multi-SpaCE incorporates Sub-SpaCE's initialization and reinitialization strategies, although, the initial mask no longer depends on the feature importance derived from an auxiliary feature-attribution method.
In summary, Multi-SpaCE introduces three major novelties: support for multivariate data, multi-objective optimization, and new mutation operators.

\subsubsection{Extension to multivariate problems}
\label{sec:multivariate_mask}

In Multi-SpaCE, the counterfactual generation process is extended to multivariate time series, where an input instance $\mathbf{x} \in \mathbb{R}^{L \times C}$ has $L$ time steps and $C$ variables (time series). The key challenge is determining the form of the binary mask of changes. We propose two alternative approaches:

\begin{itemize}
    \item \textbf{Common mask across channels:} In this approach, a single binary mask $\mathbf{m} \in \{0,1\}^L$ is shared across all channels. This means that if a change is applied at time $t$ in one channel, it is applied across all channels at the same time step. The counterfactual generation function is defined as: $x'_{ij} = f_{com}(x_{ij}|m_i, x^{nun}_{ij}) =  
    \begin{cases}
       x^{nun}_{ij},  &\text{if } m_i = 1;\\
       x_{ij},  &\text{else.} \\
     \end{cases}$,  $\forall i \in \{1, \dots, L\}, j \in \{1, \dots, C\}$.
     %, \cred{where $x^{nun}$ is obtained in the same manner as in Sub-SpaCE}.
     This approach simplifies the optimization process, reducing the number of parameters to optimize and making it agnostic to the number of channels $C$. However, it may produce suboptimal solutions, because it enforces uniform changes across all channels, which may not align with the specific dynamics of multivariate data.
    \item \textbf{Independent mask for each channel:} A more flexible approach allows us to have independent masks for each time-channel pair, represented by a binary matrix $\mathbf{M} \in \{0,1\}^{L \times C}$. There, changes are applied individually to each channel: $x'_{ij} = f_{ind}(x_{ij}|M_{ij}, x^{nun}_{ij}) = 
    \begin{cases}
       x^{nun}_{ij},  &\text{if } M_{ij} = 1;\\
       x_{ij},  &\text{else.} \\
     \end{cases}$, $\forall i \in \{1, \dots, L\}, j \in \{1, \dots, C\}$. This approach provides greater flexibility, allowing channel-specific modifications. However, the search space becomes significantly larger, potentially slowing convergence.
\end{itemize}

In both cases the NUN is the same as in Sub-SpaCE: the instance, $\mathbf{x}^{nun}$, closest to $\mathbf{x}$ in Euclidean space.
Both approaches have been tested, and the best results (see Appendix~\ref{app:ablation}) are obtained when both are used: the common mask across channels is used at the beggining of the optimization, and the independent mask is used to then improve the solution. The pseudocode of Multi-SpaCE is shown in Algorithm~\ref{algo:multisubspace}.

\begin{algorithm}
\caption{Multi-SpaCE}
\label{algo:multisubspace}
\textbf{Input:} Original time series $\mathbf{x}$, black-box classifier $b$, nearest unlike neighbor $\mathbf{x}^{nun}$, desired class $y^{nun}$, number of global channel optimization generations $G_1$, number of subsequence pruning generations $G_2$, population size $N$, initialization percentage activations $h$, increase an increase in the percentage activations $h_{inc}$, reinitialization iteration limit $G_{reinit}$, probability of extension, compression and pruning $p^{e}$, $p^{c}$, $p^{p}$

\textbf{Output:} Pareto front $\mathcal{P}_{pareto}$

\begin{algorithmic}[1]
\State $g \gets 0$
\State $\mathcal{P} \gets \text{initPopulation}(N, h)$ \Comment{as in Sub-SpaCE but purely random}
\While{$g < G_1$}
    \State $\mathcal{P}, valid \gets \text{OPTIMIZE\_OBJECTIVES}(\mathcal{P}, \mathbf{x}, b, \mathbf{x}^{nun}, y^{nun}, p^{e}, p^{c}, 0)$
    \If{$valid = 0$ and $g = G_{reinit}$}
        \State $g \gets 0$
        \State $h \gets \text{clipActivations}(h + h_{inc})$ \Comment{clip \% of activations in mask to 100\%} 
        \State $\mathcal{P} \gets \text{initPopulation}(N, h)$
    \Else
        \State $g \gets g + 1$
    \EndIf
\EndWhile
\For{generation $g = 1$ to $G_2$}
    \State $\mathcal{P} \gets \text{OPTIMIZE\_OBJECTIVES}(\mathcal{P}, \mathbf{x}, b, \mathbf{x}^{nun}, y^{nun}, 0, 0, p^{p})$
\EndFor
\State Extract best front $\mathcal{P}_{pareto}$ from $\mathcal{P}$
\State \textbf{Return} $\mathcal{P}_{pareto}$ 
\end{algorithmic}
\end{algorithm}

\subsubsection{Subsequence Mutation operators}
\label{sec:mutation}

To express the mutation operators, let $\mathcal{S}^M=\{S^{M(0)}, S^{M(1)}, ..., S^{M(k)}\}$ represent the set of subsequences in $\mathbf{M}$, where each subsequence $S^{M(k)} = (s, c, \ell)$ is defined by its start $s$, channel $c$, and length $\ell$. Any subsequence can also be represented as a binary matrix $\mathbf{S^{M(k)}} \in \{0,1\}^{L \times C}$, with ones in the position of every subsequence $\mathbf{S^{M(k)}_{ij}} = 
\begin{cases} 
    1,  &\quad (1 \leq i \leq s+\ell) \land (j = c); \\ 0,  &\quad\text{otherwise.} \\ 
\end{cases}$, $\forall i,j$. The original $\mathbf{M} = \sum_{k=1}^{K} \mathbf{S^{M(k)}}$ can be reconstructed by the union of al subsequences in $\mathcal{S}_M$ in their matrix form. 

\textbf{Subsequence Extension and Compression Mutations:} Multi-SpaCE modifies the binary mask by extending or compressing subsequences in the same way as Sub-SpaCE, but repeating the process for every channel if the independent mask $\mathbf{M}$ configuration is being used. Let $p^e$ and $p^c$ represent the probability of extending and compressing subsequences, respectively. The mutated mask $\mathbf{M'}$ can be defined as:
\vspace*{-10pt}
\begin{equation}\label{eq:extension_mutation_matrix}
    \mathbf{M'} = \sum_{k=1}^{K} \mathbf{E^{M(k)}} \oplus \mathbf{S^{M(k)}}
    \quad \text{given }
    \mathbf{E^{M(k)}_{ij}} = 
    \begin{cases}
        \delta^{e}_{ij},  & (i = s-1) \lor (i=s+\ell); \\
        0,  & \text{otherwise.} \\
    \end{cases}
\end{equation}
\vspace*{-10pt}
\begin{equation}\label{eq:compresion_mutation_matrix}
    \mathbf{M'} = \sum_{k=1}^{K} \mathbf{C^{M(k)}} \oplus \mathbf{S^{M(k)}} 
    \quad \text{given }
    \mathbf{C^{M(k)}_{ij}} = 
    \begin{cases}
        \delta^{c}_{ij},  & (i = s) \lor (i=s+\ell - 1); \\
        0,  &\text{otherwise.} \\
    \end{cases}
\end{equation}

Where $\oplus$ is the element-wise XOR operator, $\mathbf{E^{M(k)}} \in \mathbb{R}^{L \times C}$ and $\mathbf{C^{M(k)}} \in \mathbb{R}^{L \times C}$ are sparse binary matrices representing the random extension and compression mutations, respectively, and $\delta^{e}_{ij} \sim \mathcal{B}(p^{e})$ and $\delta^{c}_{ij} \sim \mathcal{B}(p^{c})$ are independently drawn Bernouilli random variables for every time step $i$ and channel $j$.

\textbf{Subsequence Removal Mutation:} An additional mutation operator is introduced to remove entire subsequences with probability $p^p$. This mutation operator improves convergence in complex multivariate settings, eliminating the need to compress any undesired subsequences incrementally, which can be inefficient when multiple or long subsequences are present. The mutated mask $\mathbf{M'}$ and the random subsequence removal matrix $\mathbf{P^{(k)}} \in \mathbb{R}^{L \times C}$ is defined as:
\vspace*{-10pt}
\begin{equation}\label{eq:removal_mutation}
    \mathbf{M'} = \sum_{k=1}^{K} \mathbf{P^{M(k)}} \oplus \mathbf{S^{M(k)}} \quad \text{given } \mathbf{P^{M(k)}_{ij}} = 
    \begin{cases}
        \delta^{p},  & s\leq i \leq s+\ell - 1; \\
        0,  &\text{otherwise.} \\
    \end{cases}
\end{equation}

Once more, $\delta^{p} \sim \mathcal{B}(p^{p})$ is a Bernoulli random variable.
%shared across all time steps of a subsequence $\mathbf{S^{M(k)}}$ for every complete subsequence.
Note that this mutation is applied at the level of entire subsequences, rather than element-wise, simplifying and accelerating the removal process for undesired subsequences.

\subsubsection{Multi-Objective Optimization} 
To eliminate the need for manual tuning of the utility function, Multi-SpaCE reformulates the counterfactual generation process as a multi-objective optimization problem that simultaneously optimizes several conflicting objectives:

\begin{equation}\label{eq:cf_mo_loss}
\begin{aligned}
\max_{\mathbf{M}} &\quad o_1(\mathbf{x'}, y^{nun}), o_2(\mathbf{M}), o_3(\mathbf{M}), o_4(\mathbf{x}, \mathbf{x'})\\
\textrm{where} &\quad o_1(\mathbf{x'}, y^{nun}) = p_b(\mathbf{x'}, y^{nun}) - \nu \cdot \mathds{1}_{class}(\mathbf{x'}, \mathbf{M}, y^{nun}) \\
&\quad o_2(\mathbf{M}) = - \frac{||\mathbf{M}||_0}{CL} - \nu \cdot \mathds{1}_{class}(\mathbf{x'}, \mathbf{M}, y^{nun}) \\
&\quad o_3(\mathbf{M}) = - \left( \frac{\sum_{j=1}^{C}\sum_{i=2}^{L} \mathds{1}_{ij}}{CL/2} \right)^{\gamma} - \nu \cdot \mathds{1}_{class}(\mathbf{x'}, \mathbf{M}, y^{nun}) \\
&\quad o_4(\mathbf{x}, \mathbf{x'}) = - \frac{e(\mathbf{x}, \mathbf{x'})}{e_{\max}} - \nu \cdot \mathds{1}_{class}(\mathbf{x'}, \mathbf{M}, y^{nun})\\
\end{aligned}
\end{equation}

Every objective is an adaptation of the terms in \eqref{eq:cf_mo_loss} to the multivariate setting. Furthermore, let us remark that the penalization term is applied to each objective, thus ensuring that non-valid solutions are always dominated by valid counterfactuals. The problem is solved using a custom implementation of \textit{Nondominated Sorting Genetic Algorithm II} (NSGA-II) \cite{deb2002nsga}, where the mutation operators are adapted to the multivariate setting. At each iteration, the algorithm computes the Pareto front, representing the best results for every possible weighting combination of objectives. Upon convergence, the algorithm provides the final Pareto front, enabling end-users to interactively explore it and select the counterfactual explanation or set of explanations that best aligns with their preferences. The process is described in Algorithm~\ref{algo:optimization}, which depicts a single iteration within the NSGA-II framework, using the mutation strategies outlined in Section~\ref{sec:mutation}. Standard NSGA-II functions, including \texttt{tournamentSelection()}, \texttt{singlePointCrossover()}, \texttt{nonDominatedSorting()}, and \texttt{crowdingDistanceSelection()} are used \cite{deb2002nsga}.

\begin{algorithm}[h]
\caption{OPTIMIZE\_OBJECTIVES()}
\label{algo:optimization}
\textbf{Input:} population $\mathcal{P}$, original time series $\mathbf{x}$, black-box classifier $b$, nearest unlike neighbor $\mathbf{x}^{nun}$, desired class $y^{nun}$, probability of extension mutation $p^{e}$, probability of compression mutation $p^{c}$, probability of pruning subsequences $p^{p}$

\textbf{Output:} new population $\mathcal{P}'$
\begin{algorithmic}[1]
\State $\mathcal{P}_{parents} \gets \text{tournamentSelection}(\mathcal{P})$
\State $\mathcal{P}_{offspring} \gets \text{crossover}(\mathcal{P}_{parents})$
\For{each offspring $\mathbf{M_n} \in \mathcal{P}_{offspring}$}
    \State Identify the set of contiguous subsequences on $M_n$, $\mathcal{S}^{M_n(k)}$

    %\State Compute $\mathbf{E^{(k)}} \forall \mathbf{S^{(k)}}$ using \eqref{eq:extension_mutation_matrix}
    \State $\mathbf{M_n} \gets \sum_{k=1}^{K} \mathbf{E^{M_n(k)}} \oplus \mathbf{S^{M_n(k)}}$

    \State Identify the set of contiguous subsequences on $M_n$, $\mathcal{S}^{M_n(k)}$
    %\State Compute $\mathbf{C^{(k)}} \forall \mathbf{S^{(k)}}$ using \eqref{eq:extension_mutation_matrix}
    \State $\mathbf{M_n} \gets \sum_{k=1}^{K} \mathbf{C^{M_n(k)}} \oplus \mathbf{S^{M_n(k)}}$

    \State Identify the set of contiguous subsequences on $M_n$, $\mathcal{S}^{M_n(k)}$
    %\State Compute $\mathbf{R^{(k)}} \forall \mathbf{S^{(k)}}$ using \eqref{eq:removal_mutation}
    \State $\mathbf{M_n} \gets \sum_{k=1}^{K} \mathbf{R^{M_n(k)}} \oplus \mathbf{S^{M_n(k)}}$
\EndFor

\State $\mathcal{P}_{ext} = \mathcal{P} \cup \mathcal{P}_{offspring}$
\For{each individual $\mathbf{M_n}$ in $\mathcal{P}$}
    \State $x'_n \gets f_{ind}(\mathbf{x}|\mathbf{M_n}, \mathbf{x^{nun}})$
    \State $Fitness_n \gets o_1(\mathbf{x'_n}, y^{nun}), o_2(\mathbf{M_n}), 
    o_3(\mathbf{M_n}), o_4(\mathbf{x}, \mathbf{x'_n})$
\EndFor
\State $\mathcal{F} \gets \text{nonDominatedSorting}(\mathcal{P}_{ext}, Fitness)$
\State $\mathcal{P}' \gets \text{crowdingDistanceSelection}(\mathcal{F}, N)$
\State \textbf{Return} New population $\mathcal{P}'$ 
\end{algorithmic}
\end{algorithm}

\section{Experiments}\label{sec:experiments}

\subsection{Setup}\label{sec:setup}
For each dataset in Table~\ref{tab:datasets}, we trained an InceptionTime \cite{ismail2020inceptiontime}, a state-of-the-art model to perform classification with time series data. Additionally, we trained several Autoencoders ($f_{AE}$) to assess the plausibility of the generated counterfactuals, as well as two alternative models: Isolation Forests ($f_{IF}$), and Local Outlier Factor ($f_{LOF}$) as used in \cite{bahri2024discord}. Details about the trained models are provided in \textit{the Supplementary Material} at \url{https://github.com/MarioRefoyo/Multi-SpaCE}. We then selected the best-performing models of each family. For the Autoencoder, the one with the lowest reconstruction error was chosen. Models with the best silhouette scores were used for the Isolation Forest and Local Outlier Factor. Classifiers and autoencoders were implemented using the TensorFlow framework, while Isolation Forests and Local Outlier Factors were trained using the Scikit-Learn library.

The parameter values of Multi-SpaCE were empirically set and kept consistent across datasets to ensure robustness in multiple scenarios. Instead of extensive fine-tuning of all possible parameters, we focused on evaluating the influence of mutation parameters and the different types of masks proposed in Section~\ref{sec:multivariate_mask}, as they are central to our framework. Based on these evaluations (see~\ref{app:ablation}), the final Multi-SpaCE uses a population size $N=100$ and a total number of generations $G=100$, with the first $G_1=75$ generations used to optimize a common mask across channels with $p^{e} = p^{c} = 0.75$ and $p^{p} = 0$, and the last $G_2=25$ generations used to prune unnecessary subsequences with the independent mask setting with $p^{p}=0.75$. We use an initialization percentage $h=20\%$, increased by $h_{inc}=20\%$ if no valid counterfactual is found after $G_{reinit}=50$ generations. The penalization term for incorrect classes was set to $\nu=100$ to strictly enforce valid counterfactuals. The implementation of Multi-SpaCE and experimental code is publicly available at \url{https://github.com/MarioRefoyo/Multi-SpaCE}.

For evaluation, a random subsample of 100 test samples was selected from each dataset, using a fixed random seed. Counterfactuals were generated for all baseline methods presented in Section~\ref{sec:baseline_methods}, and the metrics described in Section~\ref{sec:metrics} were computed. 

Unlike the baseline methods, which return a single counterfactual considered optimal according to their specific criteria, Multi-SpaCE generates a set of solutions that are optimal under different weightings of the objectives in \eqref{eq:cf_mo_loss}. To compare Multi-SpaCE with the baselines, we use the utility function proposed in Sub-SpaCE \cite{refoyo2024subspace} to assign weights to each objective: a weight of 0.1 to the adversarial objective ($o_1$), 0.3 to sparsity ($o_2$), 0.4 to the number of subsequences ($o_3$), and 0.2 to plausibility ($o_4$). After applying the utility function, we select the counterfactual from the Multi-SpaCE solution set that maximizes the utility score. This selected counterfactual is then used to compute the metrics, with the results reported in Section~\ref{sec:results}.

\subsection{Datasets}\label{sec:datasets}
The performance of Multi-SpaCE was evaluated on datasets from the UCR archive \cite{UCRArchive} and the UEA archive \cite{UEAArchive}. Only datasets where the classifier achieved an F1-score greater than 80\% were considered, ensuring that counterfactuals were generated for reasonably accurate models.

Under the multivariate setting, only 10 datasets surpassed the 80\% threshold. For the univariate setting, over 40 datasets met the criterion, out of which 15 were selected to ensure diversity based on the following criteria: i) number of classes; ii) time series length; and iii) different data characteristics. In this regard, half of the chosen datasets had a binary classification outcome, as some baseline methods are tailored for binary classification. Additionally, we selected extreme cases, such as \textit{NonInvasiveFatalECGThorax2}, which has a large number of classes (42 classes). Datasets with varying lengths were included, such as \textit{ItalyPowerDemand} (24 steps) and \textit{HandOutlines} (2709 steps). Finally, we selected datasets with diverse characteristics. For instance, \textit{FordA} was chosen since frequencies are representative of output classes,
while datasets like \textit{ECG200} and \textit{TwoPatterns} were included as waveform patterns are what define class separability.
Table~\ref{tab:datasets} provides a detailed listing of all datasets (train and test size, time series length, number of channels and classes), as well as the test F1-scores of the trained classifiers.

\mycomment{
\begin{itemize}
    \item \textbf{Number of Classes:} Half of the datasets had a binary classification outcome, as some baseline methods are tailored for binary classification. Additionally, we selected extreme cases, such as \textit{NonInvasiveFatalECGThorax2}, with a high number of classes.
    
    \item \textbf{Time Series Length:} Datasets with varying lengths were included, such as \textit{ItalyPowerDemand} (24 steps) and \textit{HandOutlines} (2709 steps).
    
    \item \textbf{Data Characteristics:} We selected datasets with diverse characteristics. For instance, \textit{FordA} was chosen due to frequencies being representative of output classes, while datasets like \textit{ECG200} and \textit{TwoPatterns} were included as waveform patterns is what define class separability.
\end{itemize}
}

\begin{center}
\setlength{\tabcolsep}{3pt}
\setlength\extrarowheight{-3pt}
\begin{table}[ht]
\centering

\resizebox{0.9\textwidth}{!}{
    \begin{tabular}{l|cccccc}
    \toprule
    Dataset & Train size & Test size & Length & Channels & Classes & F1-score \\
    \midrule
    Coffee & 28 & 28 & 1 & 286 & 2 & 1.000 \\
    Gunpoint & 50 & 150 & 1 & 150 & 2 & 1.000 \\
    Plane & 105 & 105 & 1 & 144 & 7 & 1.000 \\
    TwoPatterns & 1000 & 4000 & 1 & 128 & 4 & 1.000 \\
    CBF & 30 & 900 & 1 & 128 & 3 & 0.999 \\
    Strawberry & 613 & 370 & 1 & 235 & 2 & 0.981 \\
    FordA & 3601 & 1320 & 1 & 500 & 2 & 0.959 \\
    ItalyPowerDemand (ItalyPower) & 67 & 1029 & 1 & 24 & 2 & 0.958 \\
    FacesUCR & 200 & 2050 & 1 & 131 & 14 & 0.951 \\
    HandOutlines & 1000 & 370 & 1 & 2709 & 2 & 0.948 \\
    NonInvasiveFatalECGThorax2 (NI-ECG2) & 1800 & 1965 & 1 & 750 & 42 & 0.948 \\
    ProximalPhalanxOutlineCorrect (PPOC) & 600 & 291 & 1 & 80 & 2 & 0.913 \\
    ECG5000 & 500 & 4500 & 1 & 140 & 5 & 0.901 \\
    ECG200 & 100 & 100 & 1 & 96 & 2 & 0.876 \\
    CinCECGTorso & 40 & 1380 & 1 & 1639 & 4 & 0.856 \\
    \midrule
    BasicMotions & 40 & 40 & 6 & 100 & 4 & 1.000 \\
    PenDigits & 7494 & 3498 & 2 & 8 & 10 & 0.988 \\
    Epilepsy & 137 & 138 & 3 & 206 & 4 & 0.971 \\
    NATOPS & 180 & 180 & 24 & 51 & 6 & 0.933 \\
    UWaveGestureLibrary (UWave) & 120 & 320 & 3 & 315 & 8 & 0.883 \\
    RacketSports & 151 & 152 & 6 & 30 & 4 & 0.875 \\
    ArticularyWordRecognition (AWR) & 275 & 300 & 9 & 144 & 25 & 0.872 \\
    Cricket & 108 & 72 & 6 & 1197 & 12 & 0.870 \\
    SelfRegulationSCP1 (SR-SCP1) & 268 & 293 & 6 & 896 & 2 & 0.837 \\
    PEMS-SF & 267 & 173 & 963 & 144 & 7 & 0.813 \\
    \bottomrule
    \end{tabular}
}
\caption{Selected data sets from the UCR and UEA archives \cite{UCRArchive, UEAArchive}.}
\label{tab:datasets}
\end{table}
\end{center}

\vspace*{-20pt}
\subsection{Baseline methods}\label{sec:baseline_methods}
The proposed method, Multi-SpaCE, was evaluated against several open-source approaches recently introduced in the literature. For univariate time series datasets, the comparison included Native Guide (NG) \cite{delaney2021}, Glacier \cite{wang2024glacier} in both of the proposed configurations, perturbing the original input space (Glacier) or the latent space of the Autoencoder (GlacierAE), as well as AB-CF \cite{li2023abcf} and DiscoX \cite{bahri2024discord}. For multivariate datasets, the evaluation was conducted against COMTE \cite{ates2021}, AB-CF \cite{li2023abcf}, and DiscoX \cite{bahri2024discord}. For all methods, we used the original implementations provided by the authors and adhered to their recommended parameter settings.

\subsection{Evaluation metrics}\label{sec:metrics}
To assess the quality of the CFEs, the following metrics were computed:
\begin{itemize}
    \item \textbf{Validity}: measured as the percentage of counterfactuals that change the original output class: $\frac{1}{N} \sum_{N} \mathds{1}_{class}$, where $\mathds{1}_{class}$ is equal to 1 when the output class of the original instance and the counterfactual's class differ. 
    
    \item \textbf{Proximity}: quantified with the $\ell_2$ distance between the original sample $\mathbf{x}$ and the counterfactual $\mathbf{x'}$. 
    
    \item \textbf{Sparsity}: evaluated as the number of changes of the counterfactual, normalized by the total length of the time series: $ \frac{||\mathbf{M}||_0}{L \times C}$. A lower score is desirable.
    
    \item \textbf{Plausibility:} measured by the Outlier Score (OS) of the generated counterfactuals. As introduced in Section~\ref{sec:setup}, we use an Autoencoder, an Isolation Forest and a Local Outlier Factor. We scale the OS to lie within the range obtained in the training data set. Ideally, the OS should be close to 0.
    
    \item \textbf{Contiguity:} measured by the Number of subsequences (NoS) of changes in the counterfactual: $ \sum_{j=1}^{C} \sum_{i=2}^{L}  \mathds{1}_{ij}$, where $\mathds{1}_{ij}$ equals 1 when a new subsequence of changes is present in the channel $c$ of the counterfactual $x'$. The lower the number of subsequences, the easier it is to understand the explanation.
\end{itemize}

We also compare the execution times and present them in the Supplementary Material.
%We emphasize the relevance of validity over the rest of the metrics, as it is crucial to obtain near-perfect scores. Otherwise, the method will suffer from limited applicability to real-world problems. 

\subsection{Results}
\label{sec:results}

In this section, we present the results in terms of validity, proximity, sparsity, plausibility, and contiguity. For all metrics (except for validity), reported results are calculated exclusively on valid CFEs. The results are organized into tables specific to each metric, with separate subtables for univariate and multivariate datasets. The best performances for each dataset are highlighted in bold, the second-best results are underlined, and dashes indicate that no valid solutions were provided for that method/dataset pair. Additionally, we rank the methods for each dataset and metric and report their average rank to provide an intuitive performance comparison across different settings.
%Metrics reported are calculated only for valid solutions.

Table~\ref{tab:results_validity} shows the validity scores for all methods. As emphasized earlier, validity is a crucial requirement for real-world applicability, ensuring that counterfactuals achieve the desired outcome. Unfortunately, most current methods treat validity as a metric rather than a restriction, leading to low performance. For multivariate datasets, Multi-SpaCE is the only method that consistently provides valid counterfactuals for all instances. For univariate datasets, NG also achieves perfect validity, matching Multi-SpaCE. These results reinforce the importance of treating validity as an essential constraint, as many methods obtain results that are far from the highest score, with extremely low scores (below $0.1$) in some cases.
This highlights a fundamental limitation that may render some methods useless for some datasets where they systematically fail to obtain valid counterfactuals.

\begin{center}

\setlength{\tabcolsep}{2pt}
\setlength\extrarowheight{-3pt}
\begin{table}[ht]

\begin{subtable}[t]{.45\textwidth}
\subcaption{Multivariate Datasets}
\label{tab:sparsity_multi}
\resizebox{0.98\textwidth}{!}{ 
    \centering
    \begin{tabular}{l|cccc}
    \hline
    Dataset & COMTE & AB-CF & DiscoX & Multi-SpaCE \\
    \hline
    AWR & \underline{0.97} & 0.4 & 0.09 & \textbf{1.0} \\
    BasicMotions & \textbf{1.0} & 0.22 & 0.08 & \textbf{1.0} \\
    Cricket & \textbf{1.0} & 0.44 & - & \textbf{1.0} \\
    Epilepsy & \textbf{1.0} & 0.38 & 0.55 & \textbf{1.0} \\
    NATOPS & \underline{0.9} & 0.67 & 0.11 & \textbf{1.0} \\
    PEMS-SF & \underline{0.99} & 0.77 & - & \textbf{1.0} \\
    PenDigits & \textbf{1.0} & 0.76 & 0.88 & \textbf{1.0} \\
    RacketSports & \underline{0.98} & 0.66 & 0.25 & \textbf{1.0} \\
    SR-SCP1 & \textbf{1.0} & 0.59 & 0.55 & \textbf{1.0} \\
    UWave & \underline{0.92} & 0.75 & 0.38 & \textbf{1.0} \\
    \hline
    Average Rank & \underline{1.5} & 3.2 & 3.75 & \textbf{1.0} \\
    \hline
    \end{tabular}
}
\end{subtable}%
\begin{subtable}[t]{.55\textwidth}
\subcaption{Univariate Datasets}
\label{tab:sparsity_uni}
\resizebox{0.98\textwidth}{!}{ 
    \centering
    \begin{tabular}{l|cccccc}
    \hline
    Dataset & NG & Glacier & Glacier(AE) & AB-CF & DiscoX & Multi-SpaCE \\
    \hline
    Coffee & \textbf{1.0} & 0.75 & 0.46 & \textbf{1.0} & \textbf{1.0} & \textbf{1.0} \\
    ECG200 & \textbf{1.0} & 0.17 & 0.13 & 0.84 & 0.86 & \textbf{1.0} \\
    FordA & \textbf{1.0} & 0.2 & 0.38 & 0.98 & 0.71 & \textbf{1.0} \\
    Gunpoint & \textbf{1.0} & 0.38 & 0.3 & \textbf{1.0} & 0.91 & \textbf{1.0} \\
    HandOutlines & \textbf{1.0} & \textbf{1.0} & 0.99 & 0.87 & 0.94 & \textbf{1.0} \\
    ItalyPower & \textbf{1.0} & 0.06 & 0.13 & 0.48 & 0.39 & \textbf{1.0} \\
    PPOC & \textbf{1.0} & 0.99 & 0.57 & 0.97 & 0.92 & \textbf{1.0} \\
    Strawberry & \textbf{1.0} & 0.88 & 0.62 & \textbf{1.0} & 0.83 & \textbf{1.0} \\
    CBF & \textbf{1.0} & - & - & \textbf{1.0} & \textbf{1.0} & \textbf{1.0} \\
    CinCECGTorso & \textbf{1.0} & - & - & \textbf{1.0} & \textbf{1.0} & \textbf{1.0} \\
    TwoPatterns & \textbf{1.0} & - & - & \textbf{1.0} & \textbf{1.0} & \textbf{1.0} \\
    ECG5000 & \textbf{1.0} & - & - & 0.67 & 0.76 & \textbf{1.0} \\
    Plane & \textbf{1.0} & - & - & \textbf{1.0} & \textbf{1.0} & \textbf{1.0} \\
    FacesUCR & \textbf{1.0} & - & - & \textbf{1.0} & \textbf{1.0} & \textbf{1.0} \\
    NI-ECG2 & \textbf{1.0} & - & - & \textbf{1.0} & \textbf{1.0} & \textbf{1.0} \\
    \hline
    Average Rank & \textbf{1.0} & 4.38 & 5.5 & 2.2 & 2.67 & \textbf{1.0} \\
    \hline
    \end{tabular}
}
\end{subtable}%
\caption{Validity results.}
\label{tab:results_validity}
\end{table}
\end{center}

Proximity results are presented in Table~\ref{tab:results_proximity}. While Multi-SpaCE does not directly optimize for proximity, it achieves competitive results. For multivariate datasets, Multi-SpaCE achieves the best average rank by a small margin. In univariate datasets, Glacier is the top performer overall, frequently achieving the lowest proximity scores, followed by Multi-SpaCE ranking as the second-best method on average, and performing the best on most datasets where Glacier is not applicable.
However, note that the number of valid counterfactuals generated by Glacier is often very low, thus reducing the relevance of its good proximity score. For example, for ECG200, Glacier obtains a proximity score of 0.42 on valid counterfactuals, but 83\% of the generated counterfactuals are not valid.

\begin{center}

\setlength{\tabcolsep}{2pt}
\setlength\extrarowheight{-3pt}
\begin{table}[ht]

\begin{subtable}[t]{.45\textwidth}
\subcaption{Multivariate Datasets}
\resizebox{0.98\textwidth}{!}{ 
    \centering
    \begin{tabular}{l|cccc}
    \hline
    Dataset & COMTE & AB-CF & DiscoX & Multi-SpaCE \\
    \hline
    AWR & 25.8 & 39.43 & \textbf{10.18} & \underline{16.52} \\
    BasicMotions & 109.12 & 125.96 & \textbf{53.91} & \underline{66.71} \\
    Cricket & \underline{71.0} & 98.58 & - & \textbf{51.67} \\
    Epilepsy & 20.85 & 21.58 & \underline{14.08} & \textbf{13.04} \\
    NATOPS & \underline{8.75} & 18.77 & 8.9 & \textbf{8.38} \\
    PEMS-SF & \textbf{4.52} & 11.93 & - & \underline{5.52} \\
    PenDigits & 95.11 & 102.05 & \underline{82.73} & \textbf{43.95} \\
    RacketSports & 65.27 & 73.45 & \textbf{38.91} & \underline{50.97} \\
    SR-SCP1 & 627.08 & 1127.41 & \underline{356.33} & \textbf{285.74} \\
    UWave & 34.18 & 25.97 & \underline{17.42} & \textbf{15.2} \\
    \hline
    Average Rank & 2.7 & 3.7 & \underline{1.75} & \textbf{1.4} \\
    \hline
    \end{tabular}
}
\end{subtable}%
\begin{subtable}[t]{.55\textwidth}
\subcaption{Univariate Datasets}
\resizebox{0.98\textwidth}{!}{ 
    \centering
    \begin{tabular}{l|cccccc}
    \hline
    Dataset & NG & Glacier & Glacier(AE) & AB-CF & DiscoX & Multi-SpaCE \\
    \hline
    Coffee & 1.32 & \textbf{0.37} & 12.32 & 1.33 & 2.93 & \underline{1.2} \\
    ECG200 & 3.04 & \textbf{0.42} & \underline{1.97} & 3.38 & 4.91 & 2.56 \\
    FordA & 13.66 & \textbf{0.59} & \underline{5.6} & 16.43 & 12.45 & 7.29 \\
    Gunpoint & 2.31 & \textbf{0.43} & 4.02 & 3.07 & 4.1 & \underline{2.27} \\
    HandOutlines & 2.45 & \textbf{0.14} & \underline{0.31} & 3.44 & 6.87 & 0.57 \\
    ItalyPower & 1.65 & \textbf{0.39} & \underline{1.28} & 1.52 & 1.95 & 1.53 \\
    PPOC & 0.33 & \textbf{0.07} & 0.25 & 0.37 & 1.13 & \underline{0.19} \\
    Strawberry & 0.81 & \textbf{0.11} & \underline{0.22} & 0.8 & 1.81 & 0.37 \\
    CBF & \underline{5.85} & - & - & 7.67 & \textbf{5.71} & 5.95 \\
    CinCECGTorso & 30.55 & - & - & 30.77 & \underline{26.7} & \textbf{18.71} \\
    TwoPatterns & 6.46 & - & - & 8.84 & \textbf{5.45} & \underline{6.02} \\
    ECG5000 & \textbf{5.02} & - & - & 7.56 & 9.74 & \underline{6.08} \\
    Plane & \underline{3.86} & - & - & 5.83 & 6.7 & \textbf{3.56} \\
    FacesUCR & \underline{5.7} & - & - & 7.6 & 9.51 & \textbf{5.47} \\
    NI-ECG2 & \underline{2.21} & - & - & 3.67 & 10.59 & \textbf{1.83} \\
    \hline
    Average Rank & 3.2 & \textbf{1.0} & 3.0 & 4.0 & 4.33 & \underline{2.2} \\
    \hline
    \end{tabular}
}
\end{subtable}%
\caption{Proximity results.}
\label{tab:results_proximity}
\end{table}
\end{center}

The plausibility results exhibit greater variability, particularly for univariate datasets (see Table~\ref{tab:results_plausibility_univariate}), where no single method consistently outperforms others across all datasets and outlier detection models.
Notably, Glacier(AE) achieves the highest plausibility scores when evaluated using the Autoencoder and Local Outlier Factor (LOF), but its performance drops to fourth place under Isolation Forest evaluation. Multi-SpaCE, maintains mid-range positions across evaluation models: it ranks second under Isolation Forest, fourth under the Autoencoder, and fifth under Local Outlier Factor.
However, let us remark again that most of the methods beating Multi-SpaCE in some datasets (e.g., Glacier, GlacierAE, DiscoX and AB-CF) often achieve very low validity scores. These results indicate Multi-SpaCE's ability to generate plausible counterfactuals across various scenarios while maintaining perfect validity, unlike most of the competing approaches.

\begin{center}
\setlength{\tabcolsep}{2pt}
\setlength\extrarowheight{-3pt}
\begin{table}[ht]
\centering
\begin{subtable}[t]{1\textwidth}
\resizebox{1\textwidth}{!}{ 
    \centering
    \begin{tabular}{l|cccccc|cccccc}
    \multirow{2}{*}{Dataset} & \multicolumn{6}{c}{OS(AE)} & \multicolumn{6}{c}{OS(IF)} \\
    & NG & Glacier & Glacier(AE) & AB-CF & DiscoX & Multi-SpaCE & NG & Glacier & Glacier(AE) & AB-CF & DiscoX & Multi-SpaCE \\
    \hline
    Coffee & 0.99 & 0.98 & \textbf{0.32} & 0.98 & \underline{0.96} & 0.98 & \underline{0.37} & 0.41 & 1.32 & \textbf{0.3} & 0.44 & 0.4 \\
    ECG200 & 0.81 & \underline{0.8} & \textbf{0.45} & 0.92 & 1.09 & 0.88 & \textbf{0.38} & 0.44 & 0.4 & 0.45 & 0.49 & \underline{0.39} \\
    FordA & 0.35 & \underline{0.31} & \textbf{0.1} & 0.37 & 0.4 & 0.47 & 0.52 & 0.5 & \underline{0.36} & 0.63 & \textbf{0.08} & 0.58 \\
    Gunpoint & 0.81 & \underline{0.76} & \textbf{0.58} & 0.78 & 0.85 & 0.8 & 0.21 & \textbf{0.16} & 0.35 & \underline{0.2} & 0.29 & 0.21 \\
    HandOutlines & 0.27 & \textbf{0.26} & \textbf{0.26} & 0.32 & 0.48 & 0.3 & \underline{0.12} & \underline{0.12} & \underline{0.12} & \textbf{0.11} & 0.17 & \underline{0.12} \\
    ItalyPower & \underline{0.55} & 0.77 & \textbf{0.26} & 0.62 & 0.96 & 0.61 & 0.23 & 0.41 & \textbf{0.22} & 0.23 & 0.32 & \textbf{0.22} \\
    PPOC & \underline{0.11} & 0.12 & \textbf{0.03} & 0.12 & 0.38 & 0.12 & \underline{0.16} & 0.19 & \textbf{0.11} & \underline{0.16} & 0.28 & 0.18 \\
    Strawberry & 0.32 & \underline{0.29} & \textbf{0.15} & 0.33 & 0.75 & 0.36 & 0.13 & 0.14 & \textbf{0.08} & \underline{0.12} & 0.18 & 0.15 \\
    CBF & 0.96 & - & - & 0.95 & \textbf{0.87} & \underline{0.9} & 0.5 & - & - & \underline{0.47} & 0.52 & \textbf{0.43} \\
    CinCECGTorso & \underline{0.79} & - & - & 0.82 & \textbf{0.63} & 0.8 & 0.41 & - & - & 0.47 & \textbf{0.31} & \underline{0.4} \\
    TwoPatterns & \underline{0.7} & - & - & 0.76 & 1.03 & \textbf{0.67} & 0.38 & - & - & \textbf{0.34} & 0.51 & \textbf{0.34} \\
    ECG5000 & 0.48 & - & - & \underline{0.42} & 0.57 & \textbf{0.4} & \underline{0.28} & - & - & 0.37 & 0.36 & \textbf{0.26} \\
    Plane & \textbf{0.51} & - & - & 0.74 & 1.04 & \underline{0.55} & \textbf{0.26} & - & - & \textbf{0.26} & 0.45 & \textbf{0.26} \\
    FacesUCR & \textbf{0.51} & - & - & 0.61 & 0.68 & \underline{0.53} & \textbf{0.41} & - & - & 0.48 & 0.52 & \textbf{0.41} \\
    NI-ECG2 & \textbf{0.1} & - & - & \textbf{0.1} & 0.15 & 0.11 & \textbf{0.19} & - & - & \textbf{0.19} & 0.33 & 0.2 \\
    \hline
    Average Rank & 2.73 & \underline{2.5} & \textbf{1.0} & 3.33 & 4.33 & 3.07 & \textbf{2.27} & 3.62 & 2.75 & 2.53 & 4.27 & \underline{2.33} \\
    \hline
    \end{tabular}
}
\end{subtable}
\vspace*{6pt}

\begin{subtable}[t]{0.55\textwidth}
\resizebox{1\textwidth}{!}{ 
    \centering
    \begin{tabular}{l|cccccc}
    \multirow{2}{*}{Dataset} & \multicolumn{6}{c}{OS(LOF)} \\
     & NG & Glacier & Glacier(AE) & AB-CF & DiscoX & Multi-SpaCE \\
    \hline
    Coffee & 0.44 & \underline{0.38} & 27.42 & \textbf{0.3} & 3.99 & 0.46 \\
    ECG200 & 0.44 & \underline{0.31} & \textbf{0.2} & 0.5 & 1.0 & 0.49 \\
    FordA & 0.3 & 0.3 & \underline{0.29} & \underline{0.29} & \textbf{0.25} & 0.39 \\
    Gunpoint & 0.11 & \textbf{0.08} & 0.11 & \underline{0.1} & 0.2 & 0.11 \\
    HandOutlines & \textbf{0.03} & \textbf{0.03} & \textbf{0.03} & \textbf{0.03} & 0.1 & \textbf{0.03} \\
    ItalyPower & 0.11 & 0.3 & \textbf{0.08} & \underline{0.09} & 0.14 & 0.1 \\
    PPOC & \underline{0.06} & \underline{0.06} & \textbf{0.04} & \underline{0.06} & 0.53 & 0.07 \\
    Strawberry & 0.11 & \underline{0.09} & \textbf{0.08} & \underline{0.09} & 0.37 & 0.11 \\
    CBF & 0.18 & - & - & \textbf{0.13} & 0.2 & \textbf{0.13} \\
    CinCECGTorso & 1.49 & - & - & 2.68 & \underline{0.81} & \textbf{0.74} \\
    TwoPatterns & 0.34 & - & - & \textbf{0.26} & 0.6 & \underline{0.3} \\
    ECG5000 & \textbf{0.24} & - & - & 0.26 & 0.48 & \underline{0.25} \\
    Plane & \textbf{0.07} & - & - & 0.12 & 0.3 & \textbf{0.07} \\
    FacesUCR & \underline{0.24} & - & - & 0.31 & 0.4 & \textbf{0.23} \\
    NI-ECG2 & \textbf{0.02} & - & - & 0.03 & 0.2 & \textbf{0.02} \\
    \hline
    Average Rank & 2.53 & 2.5 & \textbf{2.0} & \underline{2.33} & 4.47 & 2.6 \\
    \hline
    \end{tabular}
}
\end{subtable}

\caption{Outlier Score (OS) results in univariate datasets. The lower the OS the higher the plausibility.}
\label{tab:results_plausibility_univariate}
\end{table}
\end{center}

\mycomment{
\begin{center}
\setlength{\tabcolsep}{2pt}
\begin{table}[ht]
\begin{subtable}[t]{1\textwidth}
\subcaption{Multivariate Datasets}
\resizebox{0.98\textwidth}{!}{ 
    \centering
    \begin{tabular}{l|cccc|cccc|cccc}
    \hline
    \multirow{2}{*}{Dataset} & \multicolumn{4}{c}{OS(AE)} & \multicolumn{4}{c}{OS(IF)} & \multicolumn{4}{c}{OS(LOF)} \\
     & COMTE & AB-CF & DiscoX & Multi-SpaCE & COMTE & AB-CF & DiscoX & Multi-SpaCE & COMTE & AB-CF & DiscoX & Multi-SpaCE \\
    \hline
    AWR & 0.78 & 0.74 & \textbf{0.67} & \underline{0.7} & \textbf{0.47} & \underline{0.51} & 0.54 & \underline{0.51} & \textbf{0.65} & \underline{0.68} & 0.73 & 0.7 \\
    BasicMotions & 0.49 & \underline{0.4} & 0.43 & \textbf{0.34} & \textbf{0.61} & \underline{0.64} & 0.73 & 0.72 & \textbf{0.78} & \underline{0.86} & 0.94 & 0.94 \\
    Cricket & \underline{0.75} & \textbf{0.74} & - & 0.76 & \textbf{0.6} & \underline{0.65} & - & 0.67 & \textbf{0.75} & \underline{0.82} & - & 0.86 \\
    Epilepsy & \textbf{0.48} & 0.59 & \underline{0.58} & 0.61 & 0.76 & \underline{0.69} & 0.7 & \textbf{0.66} & 0.93 & 0.88 & \underline{0.86} & \textbf{0.84} \\
    NATOPS & 0.79 & \textbf{0.62} & 0.71 & \underline{0.65} & \textbf{0.69} & 0.75 & 0.74 & \underline{0.72} & \textbf{0.76} & 0.81 & \underline{0.79} & \underline{0.79} \\
    PEMS-SF & 0.54 & \textbf{0.5} & - & \underline{0.51} & \textbf{0.77} & 0.81 & - & \underline{0.78} & \underline{0.94} & \textbf{0.92} & - & 0.95 \\
    PenDigits & \textbf{0.19} & 0.3 & 0.56 & \underline{0.27} & 0.73 & \underline{0.66} & \textbf{0.6} & 0.68 & 0.78 & \underline{0.68} & \textbf{0.37} & 0.75 \\
    RacketSports & \underline{0.74} & \underline{0.74} & 0.75 & \textbf{0.73} & 0.65 & \textbf{0.62} & \underline{0.64} & \underline{0.64} & 0.87 & \textbf{0.86} & \textbf{0.86} & \textbf{0.86} \\
    SR-SCP1 & 0.62 & \textbf{0.42} & 0.61 & \underline{0.44} & 0.81 & \underline{0.8} & 0.81 & \textbf{0.79} & 0.91 & 0.9 & \textbf{0.84} & \underline{0.89} \\
    UWave & \textbf{0.45} & 0.72 & 0.72 & \underline{0.62} & 0.63 & 0.57 & \underline{0.54} & \textbf{0.52} & 0.61 & \textbf{0.58} & 0.6 & \textbf{0.58} \\
    \hline
    Average Rank & 2.6 & \textbf{2.0} & 2.88 & \underline{2.1} & 2.4 & \underline{2.3} & 2.75 & \textbf{2.0} & 2.6 & \textbf{2.1} & \underline{2.12} & 2.2 \\
    \hline
    \end{tabular}
}
\end{subtable}

\begin{subtable}[t]{1\textwidth}
\subcaption{Univariate Datasets}
\resizebox{0.98\textwidth}{!}{ 
    \centering
    \begin{tabular}{l|cccccc|cccccc|cccccc}
    \hline
    \multirow{2}{*}{Dataset} & \multicolumn{6}{c}{OS(AE)} & \multicolumn{6}{c}{OS(IF)} & \multicolumn{6}{c}{OS(LOF)} \\
     & NG & Glacier & Glacier(AE) & AB-CF & DiscoX & Multi-SpaCE & NG & Glacier & Glacier(AE) & AB-CF & DiscoX & Multi-SpaCE & NG & Glacier & Glacier(AE) & AB-CF & DiscoX & Multi-SpaCE \\
    \hline
    Coffee & 0.98 & 0.98 & \textbf{0.32} & 0.98 & \underline{0.96} & 0.98 & 0.68 & \underline{0.58} & \textbf{-0.32} & 0.69 & 0.6 & 0.63 & 0.71 & 0.62 & \textbf{-25.48} & 0.74 & \underline{-3.86} & 0.6 \\
    ECG200 & \underline{0.69} & 0.84 & \textbf{0.39} & 0.85 & 1.02 & 0.86 & 0.67 & 0.66 & 0.82 & 0.68 & \textbf{0.6} & \underline{0.63} & 0.74 & 0.8 & 0.82 & \underline{0.53} & \textbf{0.06} & 0.66 \\
    FordA & \underline{0.35} & 0.39 & \textbf{0.1} & 0.38 & 0.4 & 0.47 & 0.53 & \underline{0.46} & 0.64 & 0.47 & 0.88 & \textbf{0.45} & 0.74 & \underline{0.65} & 0.7 & 0.73 & 0.75 & \textbf{0.6} \\
    Gunpoint & \underline{0.77} & 0.78 & \textbf{0.56} & 0.81 & 0.86 & 0.79 & 0.8 & 0.81 & \textbf{0.68} & 0.8 & \underline{0.75} & 0.8 & \underline{0.9} & 0.91 & 0.92 & 0.91 & \textbf{0.84} & \underline{0.9} \\
    HandOutlines & \textbf{0.29} & 0.95 & 0.35 & \underline{0.33} & 0.48 & 0.48 & \underline{0.88} & 0.89 & 0.89 & \underline{0.88} & \textbf{0.83} & \underline{0.88} & \underline{0.97} & 0.98 & 0.98 & \underline{0.97} & \textbf{0.89} & \underline{0.97} \\
    ItalyPower & \underline{0.61} & 0.63 & \textbf{0.27} & 0.69 & 0.97 & 0.65 & 0.76 & \textbf{0.68} & 0.74 & 0.75 & \underline{0.7} & 0.76 & 0.89 & \textbf{0.81} & 0.89 & 0.9 & \underline{0.86} & 0.89 \\
    PPOC & 0.14 & 0.13 & \textbf{0.03} & \underline{0.08} & 0.52 & 0.16 & 0.84 & 0.81 & 0.93 & 0.9 & \textbf{0.67} & \underline{0.8} & \underline{0.82} & 0.93 & 0.96 & 0.96 & \textbf{-0.17} & 0.9 \\
    Strawberry & 0.35 & \underline{0.28} & \textbf{0.18} & 0.41 & 0.74 & 0.32 & 0.85 & \underline{0.84} & 0.87 & 0.88 & \textbf{0.79} & 0.85 & \underline{0.89} & \underline{0.89} & \underline{0.89} & 0.9 & \textbf{0.4} & \underline{0.89} \\
    CBF & \underline{0.93} & - & - & 0.99 & \textbf{0.87} & \underline{0.93} & 0.58 & - & - & \textbf{0.49} & 0.55 & \underline{0.54} & \textbf{0.84} & - & - & \underline{0.86} & \underline{0.86} & \underline{0.86} \\
    CinCECGTorso & 0.86 & - & - & \underline{0.75} & \textbf{0.6} & 0.77 & \textbf{0.58} & - & - & \underline{0.6} & 0.67 & 0.63 & \textbf{-3.25} & - & - & \underline{-0.92} & -0.09 & -0.29 \\
    TwoPatterns & \underline{0.74} & - & - & 0.75 & 1.03 & \textbf{0.65} & 0.64 & - & - & \underline{0.63} & \textbf{0.5} & 0.65 & 0.71 & - & - & 0.69 & \textbf{0.41} & \underline{0.67} \\
    ECG5000 & \underline{0.39} & - & - & 0.54 & 0.55 & \textbf{0.38} & 0.68 & - & - & \textbf{0.57} & \underline{0.64} & 0.74 & 0.77 & - & - & \underline{0.66} & \textbf{0.57} & 0.75 \\
    Plane & \textbf{0.5} & - & - & 0.79 & 1.09 & \underline{0.54} & 0.76 & - & - & \underline{0.72} & \textbf{0.54} & 0.75 & 0.93 & - & - & \underline{0.87} & \textbf{0.63} & 0.91 \\
    FacesUCR & \textbf{0.53} & - & - & 0.6 & 0.65 & \underline{0.55} & 0.57 & - & - & \textbf{0.53} & \underline{0.54} & 0.55 & 0.73 & - & - & \textbf{0.71} & \underline{0.72} & 0.73 \\
    NI-ECG2 & \textbf{0.1} & - & - & \textbf{0.1} & 0.13 & 0.12 & 0.82 & - & - & 0.83 & \textbf{0.71} & \underline{0.8} & 0.98 & - & - & \underline{0.97} & \textbf{0.87} & \underline{0.97} \\
    \hline
    Average Rank & \underline{2.2} & 3.38 & \textbf{1.25} & 3.2 & 4.2 & 3.2 & 3.47 & 3.0 & 4.0 & 3.13 & \textbf{2.07} & \underline{2.87} & 3.07 & 3.38 & 3.88 & 3.27 & \textbf{1.8} & \underline{2.47} \\
    \hline
    \end{tabular}
}
\end{subtable}

\caption{Plausibility (OS) results.}
\end{table}
\end{center}

\begin{center}
\setlength{\tabcolsep}{2pt}
\begin{table}[ht]
\begin{subtable}[t]{1\textwidth}
\subcaption{Multivariate Datasets}
\resizebox{0.98\textwidth}{!}{ 
    \centering
    \begin{tabular}{l|cccc|cccc|cccc}
    \hline
    \multirow{2}{*}{Dataset} & \multicolumn{4}{c}{OS(AE)} & \multicolumn{4}{c}{OS(IF)} & \multicolumn{4}{c}{OS(LOF)} \\
     & COMTE & AB-CF & DiscoX & Multi-SpaCE & COMTE & AB-CF & DiscoX & Multi-SpaCE & COMTE & AB-CF & DiscoX & Multi-SpaCE \\
    \hline
    AWR & 0.78 & 0.74 & \textbf{0.67} & \underline{0.7} & \textbf{0.47} & \underline{0.51} & 0.54 & \underline{0.51} & \textbf{0.65} & \underline{0.68} & 0.73 & 0.7 \\
    BasicMotions & 0.49 & \underline{0.4} & 0.43 & \textbf{0.34} & \textbf{0.61} & \underline{0.64} & 0.73 & 0.72 & \textbf{0.78} & \underline{0.86} & 0.94 & 0.94 \\
    Cricket & \underline{0.75} & \textbf{0.74} & - & 0.76 & \textbf{0.6} & \underline{0.65} & - & 0.67 & \textbf{0.75} & \underline{0.82} & - & 0.86 \\
    Epilepsy & \textbf{0.48} & 0.59 & \underline{0.58} & 0.61 & 0.76 & \underline{0.69} & 0.7 & \textbf{0.66} & 0.93 & 0.88 & \underline{0.86} & \textbf{0.84} \\
    NATOPS & 0.79 & \textbf{0.62} & 0.71 & \underline{0.65} & \textbf{0.69} & 0.75 & 0.74 & \underline{0.72} & \textbf{0.76} & 0.81 & \underline{0.79} & \underline{0.79} \\
    PEMS-SF & 0.54 & \textbf{0.5} & - & \underline{0.51} & \textbf{0.77} & 0.81 & - & \underline{0.78} & \underline{0.94} & \textbf{0.92} & - & 0.95 \\
    PenDigits & \textbf{0.19} & 0.3 & 0.56 & \underline{0.27} & 0.73 & \underline{0.66} & \textbf{0.6} & 0.68 & 0.78 & \underline{0.68} & \textbf{0.37} & 0.75 \\
    RacketSports & \underline{0.74} & \underline{0.74} & 0.75 & \textbf{0.73} & 0.65 & \textbf{0.62} & \underline{0.64} & \underline{0.64} & 0.87 & \textbf{0.86} & \textbf{0.86} & \textbf{0.86} \\
    SR-SCP1 & 0.62 & \textbf{0.42} & 0.61 & \underline{0.44} & 0.81 & \underline{0.8} & 0.81 & \textbf{0.79} & 0.91 & 0.9 & \textbf{0.84} & \underline{0.89} \\
    UWave & \textbf{0.45} & 0.72 & 0.72 & \underline{0.62} & 0.63 & 0.57 & \underline{0.54} & \textbf{0.52} & 0.61 & \textbf{0.58} & 0.6 & \textbf{0.58} \\
    \hline
    Average Rank & 2.6 & \textbf{2.0} & 2.88 & \underline{2.1} & 2.4 & \underline{2.3} & 2.75 & \textbf{2.0} & 2.6 & \textbf{2.1} & \underline{2.12} & 2.2 \\
    \hline
    \end{tabular}
}
\end{subtable}

\begin{subtable}[t]{1\textwidth}
\subcaption{Univariate Datasets}
\resizebox{0.98\textwidth}{!}{ 
    \centering
    \begin{tabular}{l|cccccc|cccccc|cccccc}
    \hline
    \multirow{2}{*}{Dataset} & \multicolumn{6}{c}{OS(AE)} & \multicolumn{6}{c}{OS(IF)} & \multicolumn{6}{c}{OS(LOF)} \\
     & NG & Glacier & Glacier(AE) & AB-CF & DiscoX & Multi-SpaCE & NG & Glacier & Glacier(AE) & AB-CF & DiscoX & Multi-SpaCE & NG & Glacier & Glacier(AE) & AB-CF & DiscoX & Multi-SpaCE \\
    \hline
    Coffee & 0.98 & 0.98 & \textbf{0.32} & 0.98 & \underline{0.96} & 0.98 & 0.68 & \underline{0.58} & \textbf{-0.32} & 0.69 & 0.6 & 0.63 & 0.71 & 0.62 & \textbf{-25.48} & 0.74 & \underline{-3.86} & 0.6 \\
    ECG200 & \underline{0.69} & 0.84 & \textbf{0.39} & 0.85 & 1.02 & 0.86 & 0.67 & 0.66 & 0.82 & 0.68 & \textbf{0.6} & \underline{0.63} & 0.74 & 0.8 & 0.82 & \underline{0.53} & \textbf{0.06} & 0.66 \\
    FordA & \underline{0.35} & 0.39 & \textbf{0.1} & 0.38 & 0.4 & 0.47 & 0.53 & \underline{0.46} & 0.64 & 0.47 & 0.88 & \textbf{0.45} & 0.74 & \underline{0.65} & 0.7 & 0.73 & 0.75 & \textbf{0.6} \\
    Gunpoint & \underline{0.77} & 0.78 & \textbf{0.56} & 0.81 & 0.86 & 0.79 & 0.8 & 0.81 & \textbf{0.68} & 0.8 & \underline{0.75} & 0.8 & \underline{0.9} & 0.91 & 0.92 & 0.91 & \textbf{0.84} & \underline{0.9} \\
    HandOutlines & \textbf{0.29} & 0.95 & 0.35 & \underline{0.33} & 0.48 & 0.48 & \underline{0.88} & 0.89 & 0.89 & \underline{0.88} & \textbf{0.83} & \underline{0.88} & \underline{0.97} & 0.98 & 0.98 & \underline{0.97} & \textbf{0.89} & \underline{0.97} \\
    ItalyPower & \underline{0.61} & 0.63 & \textbf{0.27} & 0.69 & 0.97 & 0.65 & 0.76 & \textbf{0.68} & 0.74 & 0.75 & \underline{0.7} & 0.76 & 0.89 & \textbf{0.81} & 0.89 & 0.9 & \underline{0.86} & 0.89 \\
    PPOC & 0.14 & 0.13 & \textbf{0.03} & \underline{0.08} & 0.52 & 0.16 & 0.84 & 0.81 & 0.93 & 0.9 & \textbf{0.67} & \underline{0.8} & \underline{0.82} & 0.93 & 0.96 & 0.96 & \textbf{-0.17} & 0.9 \\
    Strawberry & 0.35 & \underline{0.28} & \textbf{0.18} & 0.41 & 0.74 & 0.32 & 0.85 & \underline{0.84} & 0.87 & 0.88 & \textbf{0.79} & 0.85 & \underline{0.89} & \underline{0.89} & \underline{0.89} & 0.9 & \textbf{0.4} & \underline{0.89} \\
    CBF & \underline{0.93} & - & - & 0.99 & \textbf{0.87} & \underline{0.93} & 0.58 & - & - & \textbf{0.49} & 0.55 & \underline{0.54} & \textbf{0.84} & - & - & \underline{0.86} & \underline{0.86} & \underline{0.86} \\
    CinCECGTorso & 0.86 & - & - & \underline{0.75} & \textbf{0.6} & 0.77 & \textbf{0.58} & - & - & \underline{0.6} & 0.67 & 0.63 & \textbf{-3.25} & - & - & \underline{-0.92} & -0.09 & -0.29 \\
    TwoPatterns & \underline{0.74} & - & - & 0.75 & 1.03 & \textbf{0.65} & 0.64 & - & - & \underline{0.63} & \textbf{0.5} & 0.65 & 0.71 & - & - & 0.69 & \textbf{0.41} & \underline{0.67} \\
    ECG5000 & \underline{0.39} & - & - & 0.54 & 0.55 & \textbf{0.38} & 0.68 & - & - & \textbf{0.57} & \underline{0.64} & 0.74 & 0.77 & - & - & \underline{0.66} & \textbf{0.57} & 0.75 \\
    Plane & \textbf{0.5} & - & - & 0.79 & 1.09 & \underline{0.54} & 0.76 & - & - & \underline{0.72} & \textbf{0.54} & 0.75 & 0.93 & - & - & \underline{0.87} & \textbf{0.63} & 0.91 \\
    FacesUCR & \textbf{0.53} & - & - & 0.6 & 0.65 & \underline{0.55} & 0.57 & - & - & \textbf{0.53} & \underline{0.54} & 0.55 & 0.73 & - & - & \textbf{0.71} & \underline{0.72} & 0.73 \\
    NI-ECG2 & \textbf{0.1} & - & - & \textbf{0.1} & 0.13 & 0.12 & 0.82 & - & - & 0.83 & \textbf{0.71} & \underline{0.8} & 0.98 & - & - & \underline{0.97} & \textbf{0.87} & \underline{0.97} \\
    \hline
    Average Rank & \underline{2.2} & 3.38 & \textbf{1.25} & 3.2 & 4.2 & 3.2 & 3.47 & 3.0 & 4.0 & 3.13 & \textbf{2.07} & \underline{2.87} & 3.07 & 3.38 & 3.88 & 3.27 & \textbf{1.8} & \underline{2.47} \\
    \hline
    \end{tabular}
}
\end{subtable}

\caption{Plausibility (OS) results.}
\end{table}
\end{center}

\begin{center}
\setlength{\tabcolsep}{2pt}
\begin{table}[H]

\begin{subtable}[t]{0.45\textwidth}
\subcaption{Multivariate Datasets}
\resizebox{0.98\textwidth}{!}{ 
    \centering
    \begin{tabular}{l|cccc}
    \multirow{2}{*}{Dataset} & \multicolumn{4}{c}{OS(AE)} \\
     & COMTE & AB-CF & DiscoX & Multi-SpaCE \\
    \hline
    AWR & 0.78 & 0.74 & \textbf{0.67} & \underline{0.7} \\
    BasicMotions & 0.49 & \underline{0.4} & 0.43 & \textbf{0.34} \\
    Cricket & \underline{0.75} & \textbf{0.74} & - & 0.76 \\
    Epilepsy & \textbf{0.48} & 0.59 & \underline{0.58} & 0.61 \\
    NATOPS & 0.79 & \textbf{0.62} & 0.71 & \underline{0.65} \\
    PEMS-SF & 0.54 & \textbf{0.5} & - & \underline{0.51} \\
    PenDigits & \textbf{0.19} & 0.3 & 0.56 & \underline{0.27} \\
    RacketSports & \underline{0.74} & \underline{0.74} & 0.75 & \textbf{0.73} \\
    SR-SCP1 & 0.62 & \textbf{0.42} & 0.61 & \underline{0.44} \\
    UWave & \textbf{0.45} & 0.72 & 0.72 & \underline{0.62} \\
    \hline
    Average Rank & 2.6 & \textbf{2.0} & 2.88 & \underline{2.1} \\
    \hline
    \end{tabular}
}
\end{subtable}%
\begin{subtable}[t]{0.55\textwidth}
\subcaption{Univariate Datasets}
\resizebox{0.98\textwidth}{!}{ 
    \centering
    \begin{tabular}{l|cccccc}
    \multirow{2}{*}{Dataset} & \multicolumn{6}{c}{OS(AE)} \\
     & NG & Glacier & Glacier(AE) & AB-CF & DiscoX & Multi-SpaCE \\
    \hline
    Coffee & 0.98 & 0.98 & \textbf{0.32} & 0.98 & \underline{0.96} & 0.98 \\
    ECG200 & \underline{0.69} & 0.84 & \textbf{0.39} & 0.85 & 1.02 & 0.86 \\
    FordA & \underline{0.35} & 0.39 & \textbf{0.1} & 0.38 & 0.4 & 0.47 \\
    Gunpoint & \underline{0.77} & 0.78 & \textbf{0.56} & 0.81 & 0.86 & 0.79 \\
    HandOutlines & \textbf{0.29} & 0.95 & 0.35 & \underline{0.33} & 0.48 & 0.48 \\
    ItalyPower & \underline{0.61} & 0.63 & \textbf{0.27} & 0.69 & 0.97 & 0.65 \\
    PPOC & 0.14 & 0.13 & \textbf{0.03} & \underline{0.08} & 0.52 & 0.16 \\
    Strawberry & 0.35 & \underline{0.28} & \textbf{0.18} & 0.41 & 0.74 & 0.32 \\
    CBF & \underline{0.93} & - & - & 0.99 & \textbf{0.87} & \underline{0.93} \\
    CinCECGTorso & 0.86 & - & - & \underline{0.75} & \textbf{0.6} & 0.77 \\
    TwoPatterns & \underline{0.74} & - & - & 0.75 & 1.03 & \textbf{0.65} \\
    ECG5000 & \underline{0.39} & - & - & 0.54 & 0.55 & \textbf{0.38} \\
    Plane & \textbf{0.5} & - & - & 0.79 & 1.09 & \underline{0.54} \\
    FacesUCR & \textbf{0.53} & - & - & 0.6 & 0.65 & \underline{0.55} \\
    NI-ECG2 & \textbf{0.1} & - & - & \textbf{0.1} & 0.13 & 0.12 \\
    \hline
    Average Rank & \underline{2.2} & 3.38 & \textbf{1.25} & 3.2 & 4.2 & 3.2 \\
    \hline
    \end{tabular}
}
\end{subtable}

\begin{subtable}[t]{0.45\textwidth}
\resizebox{0.98\textwidth}{!}{ 
    \centering
    \begin{tabular}{l|cccc}
    \multirow{2}{*}{Dataset} & \multicolumn{4}{c}{OS(IF)} \\
     & COMTE & AB-CF & DiscoX & Multi-SpaCE \\
    \hline
    AWR & \textbf{0.47} & \underline{0.51} & 0.54 & \underline{0.51} \\
    BasicMotions & \textbf{0.61} & \underline{0.64} & 0.73 & 0.72 \\
    Cricket & \textbf{0.6} & \underline{0.65} & - & 0.67 \\
    Epilepsy & 0.76 & \underline{0.69} & 0.7 & \textbf{0.66} \\
    NATOPS & \textbf{0.69} & 0.75 & 0.74 & \underline{0.72} \\
    PEMS-SF & \textbf{0.77} & 0.81 & - & \underline{0.78} \\
    PenDigits & 0.73 & \underline{0.66} & \textbf{0.6} & 0.68 \\
    RacketSports & 0.65 & \textbf{0.62} & \underline{0.64} & \underline{0.64} \\
    SR-SCP1 & 0.81 & \underline{0.8} & 0.81 & \textbf{0.79} \\
    UWave & 0.63 & 0.57 & \underline{0.54} & \textbf{0.52} \\
    \hline
    Average Rank & 2.4 & \underline{2.3} & 2.75 & \textbf{2.0} \\
    \hline
    \end{tabular}
}
\end{subtable}%
\begin{subtable}[t]{0.55\textwidth}
\resizebox{0.98\textwidth}{!}{ 
    \centering
    \begin{tabular}{l|cccccc}
    \multirow{2}{*}{Dataset} & \multicolumn{6}{c}{OS(IF)} \\
     & NG & Glacier & Glacier(AE) & AB-CF & DiscoX & Multi-SpaCE \\
    \hline
    Coffee & 0.68 & \underline{0.58} & \textbf{-0.32} & 0.69 & 0.6 & 0.63 \\
    ECG200 & 0.67 & 0.66 & 0.82 & 0.68 & \textbf{0.6} & \underline{0.63} \\
    FordA & 0.53 & \underline{0.46} & 0.64 & 0.47 & 0.88 & \textbf{0.45} \\
    Gunpoint & 0.8 & 0.81 & \textbf{0.68} & 0.8 & \underline{0.75} & 0.8 \\
    HandOutlines & \underline{0.88} & 0.89 & 0.89 & \underline{0.88} & \textbf{0.83} & \underline{0.88} \\
    ItalyPower & 0.76 & \textbf{0.68} & 0.74 & 0.75 & \underline{0.7} & 0.76 \\
    PPOC & 0.84 & 0.81 & 0.93 & 0.9 & \textbf{0.67} & \underline{0.8} \\
    Strawberry & 0.85 & \underline{0.84} & 0.87 & 0.88 & \textbf{0.79} & 0.85 \\
    CBF & 0.58 & - & - & \textbf{0.49} & 0.55 & \underline{0.54} \\
    CinCECGTorso & \textbf{0.58} & - & - & \underline{0.6} & 0.67 & 0.63 \\
    TwoPatterns & 0.64 & - & - & \underline{0.63} & \textbf{0.5} & 0.65 \\
    ECG5000 & 0.68 & - & - & \textbf{0.57} & \underline{0.64} & 0.74 \\
    Plane & 0.76 & - & - & \underline{0.72} & \textbf{0.54} & 0.75 \\
    FacesUCR & 0.57 & - & - & \textbf{0.53} & \underline{0.54} & 0.55 \\
    NI-ECG2 & 0.82 & - & - & 0.83 & \textbf{0.71} & \underline{0.8} \\
    \hline
    Average Rank & 3.47 & 3.0 & 4.0 & 3.13 & \textbf{2.07} & \underline{2.87} \\
    \hline
    \end{tabular}
}
\end{subtable}%

\begin{subtable}[t]{0.45\textwidth}
\resizebox{0.98\textwidth}{!}{ 
    \centering
    \begin{tabular}{l|cccc}
    \multirow{2}{*}{Dataset} & \multicolumn{4}{c}{OS(LOF)} \\
     & COMTE & AB-CF & DiscoX & Multi-SpaCE \\
    \hline
    AWR & \textbf{0.65} & \underline{0.68} & 0.73 & 0.7 \\
    BasicMotions & \textbf{0.78} & \underline{0.86} & 0.94 & 0.94 \\
    Cricket & \textbf{0.75} & \underline{0.82} & - & 0.86 \\
    Epilepsy & 0.93 & 0.88 & \underline{0.86} & \textbf{0.84} \\
    NATOPS & \textbf{0.76} & 0.81 & \underline{0.79} & \underline{0.79} \\
    PEMS-SF & \underline{0.94} & \textbf{0.92} & - & 0.95 \\
    PenDigits & 0.78 & \underline{0.68} & \textbf{0.37} & 0.75 \\
    RacketSports & 0.87 & \textbf{0.86} & \textbf{0.86} & \textbf{0.86} \\
    SR-SCP1 & 0.91 & 0.9 & \textbf{0.84} & \underline{0.89} \\
    UWave & 0.61 & \textbf{0.58} & 0.6 & \textbf{0.58} \\
    \hline
    Average Rank & 2.6 & \textbf{2.1} & \underline{2.12} & 2.2 \\
    \hline
    \end{tabular}
}
\end{subtable}%
\begin{subtable}[t]{0.55\textwidth}
\resizebox{0.98\textwidth}{!}{ 
    \centering
    \begin{tabular}{l|cccccc}
    \multirow{2}{*}{Dataset} & \multicolumn{6}{c}{OS(LOF)} \\
     & NG & Glacier & Glacier(AE) & AB-CF & DiscoX & Multi-SpaCE \\
    \hline
    Coffee & 0.71 & 0.62 & \textbf{-25.48} & 0.74 & \underline{-3.86} & 0.6 \\
    ECG200 & 0.74 & 0.8 & 0.82 & \underline{0.53} & \textbf{0.06} & 0.66 \\
    FordA & 0.74 & \underline{0.65} & 0.7 & 0.73 & 0.75 & \textbf{0.6} \\
    Gunpoint & \underline{0.9} & 0.91 & 0.92 & 0.91 & \textbf{0.84} & \underline{0.9} \\
    HandOutlines & \underline{0.97} & 0.98 & 0.98 & \underline{0.97} & \textbf{0.89} & \underline{0.97} \\
    ItalyPower & 0.89 & \textbf{0.81} & 0.89 & 0.9 & \underline{0.86} & 0.89 \\
    PPOC & \underline{0.82} & 0.93 & 0.96 & 0.96 & \textbf{-0.17} & 0.9 \\
    Strawberry & \underline{0.89} & \underline{0.89} & \underline{0.89} & 0.9 & \textbf{0.4} & \underline{0.89} \\
    CBF & \textbf{0.84} & - & - & \underline{0.86} & \underline{0.86} & \underline{0.86} \\
    CinCECGTorso & \textbf{-3.25} & - & - & \underline{-0.92} & -0.09 & -0.29 \\
    TwoPatterns & 0.71 & - & - & 0.69 & \textbf{0.41} & \underline{0.67} \\
    ECG5000 & 0.77 & - & - & \underline{0.66} & \textbf{0.57} & 0.75 \\
    Plane & 0.93 & - & - & \underline{0.87} & \textbf{0.63} & 0.91 \\
    FacesUCR & 0.73 & - & - & \textbf{0.71} & \underline{0.72} & 0.73 \\
    NI-ECG2 & 0.98 & - & - & \underline{0.97} & \textbf{0.87} & \underline{0.97} \\
    \hline
    Average Rank & 3.07 & 3.38 & 3.88 & 3.27 & \textbf{1.8} & \underline{2.47} \\
    \hline
    \end{tabular}
}
\end{subtable}%

\caption{Plausibility (OS) results.}
\end{table}
\end{center}

\begin{center}
\setlength{\tabcolsep}{2pt}
\begin{table}[ht]

\begin{minipage}{0.45\linewidth}
\begin{subtable}[t]{1\textwidth}
\subcaption{Multivariate Datasets}
\resizebox{0.98\textwidth}{!}{ 
    \begin{tabular}{l|cccc}
    \multirow{2}{*}{Dataset} & \multicolumn{4}{c}{OS(AE)} \\
     & COMTE & AB-CF & DiscoX & Multi-SpaCE \\
    \hline
    AWR & 0.78 & 0.74 & \textbf{0.67} & \underline{0.7} \\
    BasicMotions & 0.49 & \underline{0.4} & 0.43 & \textbf{0.34} \\
    Cricket & \underline{0.75} & \textbf{0.74} & - & 0.76 \\
    Epilepsy & \textbf{0.48} & 0.59 & \underline{0.58} & 0.61 \\
    NATOPS & 0.79 & \textbf{0.62} & 0.71 & \underline{0.65} \\
    PEMS-SF & 0.54 & \textbf{0.5} & - & \underline{0.51} \\
    PenDigits & \textbf{0.19} & 0.3 & 0.56 & \underline{0.27} \\
    RacketSports & \underline{0.74} & \underline{0.74} & 0.75 & \textbf{0.73} \\
    SR-SCP1 & 0.62 & \textbf{0.42} & 0.61 & \underline{0.44} \\
    UWave & \textbf{0.45} & 0.72 & 0.72 & \underline{0.62} \\
    \hline
    Average Rank & 2.6 & \textbf{2.0} & 2.88 & \underline{2.1} \\
    \hline
    \end{tabular}
}
\end{subtable}
\vspace*{32pt}

\begin{subtable}[t]{1\textwidth}
\resizebox{0.98\textwidth}{!}{ 
    \begin{tabular}{l|cccc}
    \multirow{2}{*}{Dataset} & \multicolumn{4}{c}{OS(IF)} \\
     & COMTE & AB-CF & DiscoX & Multi-SpaCE \\
    \hline
    AWR & \textbf{0.47} & \underline{0.51} & 0.54 & \underline{0.51} \\
    BasicMotions & \textbf{0.61} & \underline{0.64} & 0.73 & 0.72 \\
    Cricket & \textbf{0.6} & \underline{0.65} & - & 0.67 \\
    Epilepsy & 0.76 & \underline{0.69} & 0.7 & \textbf{0.66} \\
    NATOPS & \textbf{0.69} & 0.75 & 0.74 & \underline{0.72} \\
    PEMS-SF & \textbf{0.77} & 0.81 & - & \underline{0.78} \\
    PenDigits & 0.73 & \underline{0.66} & \textbf{0.6} & 0.68 \\
    RacketSports & 0.65 & \textbf{0.62} & \underline{0.64} & \underline{0.64} \\
    SR-SCP1 & 0.81 & \underline{0.8} & 0.81 & \textbf{0.79} \\
    UWave & 0.63 & 0.57 & \underline{0.54} & \textbf{0.52} \\
    \hline
    Average Rank & 2.4 & \underline{2.3} & 2.75 & \textbf{2.0} \\
    \hline
    \end{tabular}
}
\end{subtable}
\vspace*{32pt}

\begin{subtable}[t]{1\textwidth}
\resizebox{0.98\textwidth}{!}{ 
    \begin{tabular}{l|cccc}
    \multirow{2}{*}{Dataset} & \multicolumn{4}{c}{OS(LOF)} \\
     & COMTE & AB-CF & DiscoX & Multi-SpaCE \\
    \hline
    AWR & \textbf{0.65} & \underline{0.68} & 0.73 & 0.7 \\
    BasicMotions & \textbf{0.78} & \underline{0.86} & 0.94 & 0.94 \\
    Cricket & \textbf{0.75} & \underline{0.82} & - & 0.86 \\
    Epilepsy & 0.93 & 0.88 & \underline{0.86} & \textbf{0.84} \\
    NATOPS & \textbf{0.76} & 0.81 & \underline{0.79} & \underline{0.79} \\
    PEMS-SF & \underline{0.94} & \textbf{0.92} & - & 0.95 \\
    PenDigits & 0.78 & \underline{0.68} & \textbf{0.37} & 0.75 \\
    RacketSports & 0.87 & \textbf{0.86} & \textbf{0.86} & \textbf{0.86} \\
    SR-SCP1 & 0.91 & 0.9 & \textbf{0.84} & \underline{0.89} \\
    UWave & 0.61 & \textbf{0.58} & 0.6 & \textbf{0.58} \\
    \hline
    Average Rank & 2.6 & \textbf{2.1} & \underline{2.12} & 2.2 \\
    \hline
    \end{tabular}
}
\end{subtable}%
\vspace*{32pt}
\end{minipage}
\begin{minipage}{0.55\linewidth}
\begin{subtable}[t]{1\textwidth}
\subcaption{Univariate Datasets}
\resizebox{0.98\textwidth}{!}{ 
    \centering
    \begin{tabular}{l|cccccc}
    \multirow{2}{*}{Dataset} & \multicolumn{6}{c}{OS(AE)} \\
     & NG & Glacier & Glacier(AE) & AB-CF & DiscoX & Multi-SpaCE \\
    \hline
    Coffee & 0.98 & 0.98 & \textbf{0.32} & 0.98 & \underline{0.96} & 0.98 \\
    ECG200 & \underline{0.69} & 0.84 & \textbf{0.39} & 0.85 & 1.02 & 0.86 \\
    FordA & \underline{0.35} & 0.39 & \textbf{0.1} & 0.38 & 0.4 & 0.47 \\
    Gunpoint & \underline{0.77} & 0.78 & \textbf{0.56} & 0.81 & 0.86 & 0.79 \\
    HandOutlines & \textbf{0.29} & 0.95 & 0.35 & \underline{0.33} & 0.48 & 0.48 \\
    ItalyPower & \underline{0.61} & 0.63 & \textbf{0.27} & 0.69 & 0.97 & 0.65 \\
    PPOC & 0.14 & 0.13 & \textbf{0.03} & \underline{0.08} & 0.52 & 0.16 \\
    Strawberry & 0.35 & \underline{0.28} & \textbf{0.18} & 0.41 & 0.74 & 0.32 \\
    CBF & \underline{0.93} & - & - & 0.99 & \textbf{0.87} & \underline{0.93} \\
    CinCECGTorso & 0.86 & - & - & \underline{0.75} & \textbf{0.6} & 0.77 \\
    TwoPatterns & \underline{0.74} & - & - & 0.75 & 1.03 & \textbf{0.65} \\
    ECG5000 & \underline{0.39} & - & - & 0.54 & 0.55 & \textbf{0.38} \\
    Plane & \textbf{0.5} & - & - & 0.79 & 1.09 & \underline{0.54} \\
    FacesUCR & \textbf{0.53} & - & - & 0.6 & 0.65 & \underline{0.55} \\
    NI-ECG2 & \textbf{0.1} & - & - & \textbf{0.1} & 0.13 & 0.12 \\
    \hline
    Average Rank & \underline{2.2} & 3.38 & \textbf{1.25} & 3.2 & 4.2 & 3.2 \\
    \hline
    \end{tabular}
}
\end{subtable}
\vspace*{4pt}

\begin{subtable}[t]{1\textwidth}
\resizebox{0.98\textwidth}{!}{ 
    \centering
    \begin{tabular}{l|cccccc}
    \multirow{2}{*}{Dataset} & \multicolumn{6}{c}{OS(IF)} \\
     & NG & Glacier & Glacier(AE) & AB-CF & DiscoX & Multi-SpaCE \\
    \hline
    Coffee & 0.68 & \underline{0.58} & \textbf{-0.32} & 0.69 & 0.6 & 0.63 \\
    ECG200 & 0.67 & 0.66 & 0.82 & 0.68 & \textbf{0.6} & \underline{0.63} \\
    FordA & 0.53 & \underline{0.46} & 0.64 & 0.47 & 0.88 & \textbf{0.45} \\
    Gunpoint & 0.8 & 0.81 & \textbf{0.68} & 0.8 & \underline{0.75} & 0.8 \\
    HandOutlines & \underline{0.88} & 0.89 & 0.89 & \underline{0.88} & \textbf{0.83} & \underline{0.88} \\
    ItalyPower & 0.76 & \textbf{0.68} & 0.74 & 0.75 & \underline{0.7} & 0.76 \\
    PPOC & 0.84 & 0.81 & 0.93 & 0.9 & \textbf{0.67} & \underline{0.8} \\
    Strawberry & 0.85 & \underline{0.84} & 0.87 & 0.88 & \textbf{0.79} & 0.85 \\
    CBF & 0.58 & - & - & \textbf{0.49} & 0.55 & \underline{0.54} \\
    CinCECGTorso & \textbf{0.58} & - & - & \underline{0.6} & 0.67 & 0.63 \\
    TwoPatterns & 0.64 & - & - & \underline{0.63} & \textbf{0.5} & 0.65 \\
    ECG5000 & 0.68 & - & - & \textbf{0.57} & \underline{0.64} & 0.74 \\
    Plane & 0.76 & - & - & \underline{0.72} & \textbf{0.54} & 0.75 \\
    FacesUCR & 0.57 & - & - & \textbf{0.53} & \underline{0.54} & 0.55 \\
    NI-ECG2 & 0.82 & - & - & 0.83 & \textbf{0.71} & \underline{0.8} \\
    \hline
    Average Rank & 3.47 & 3.0 & 4.0 & 3.13 & \textbf{2.07} & \underline{2.87} \\
    \hline
    \end{tabular}
}
\end{subtable}
\vspace*{4pt}

\begin{subtable}[t]{1\textwidth}
\resizebox{0.98\textwidth}{!}{ 
    \centering
    \begin{tabular}{l|cccccc}
    \multirow{2}{*}{Dataset} & \multicolumn{6}{c}{OS(LOF)} \\
     & NG & Glacier & Glacier(AE) & AB-CF & DiscoX & Multi-SpaCE \\
    \hline
    Coffee & 0.71 & 0.62 & \textbf{-25.48} & 0.74 & \underline{-3.86} & 0.6 \\
    ECG200 & 0.74 & 0.8 & 0.82 & \underline{0.53} & \textbf{0.06} & 0.66 \\
    FordA & 0.74 & \underline{0.65} & 0.7 & 0.73 & 0.75 & \textbf{0.6} \\
    Gunpoint & \underline{0.9} & 0.91 & 0.92 & 0.91 & \textbf{0.84} & \underline{0.9} \\
    HandOutlines & \underline{0.97} & 0.98 & 0.98 & \underline{0.97} & \textbf{0.89} & \underline{0.97} \\
    ItalyPower & 0.89 & \textbf{0.81} & 0.89 & 0.9 & \underline{0.86} & 0.89 \\
    PPOC & \underline{0.82} & 0.93 & 0.96 & 0.96 & \textbf{-0.17} & 0.9 \\
    Strawberry & \underline{0.89} & \underline{0.89} & \underline{0.89} & 0.9 & \textbf{0.4} & \underline{0.89} \\
    CBF & \textbf{0.84} & - & - & \underline{0.86} & \underline{0.86} & \underline{0.86} \\
    CinCECGTorso & \textbf{-3.25} & - & - & \underline{-0.92} & -0.09 & -0.29 \\
    TwoPatterns & 0.71 & - & - & 0.69 & \textbf{0.41} & \underline{0.67} \\
    ECG5000 & 0.77 & - & - & \underline{0.66} & \textbf{0.57} & 0.75 \\
    Plane & 0.93 & - & - & \underline{0.87} & \textbf{0.63} & 0.91 \\
    FacesUCR & 0.73 & - & - & \textbf{0.71} & \underline{0.72} & 0.73 \\
    NI-ECG2 & 0.98 & - & - & \underline{0.97} & \textbf{0.87} & \underline{0.97} \\
    \hline
    Average Rank & 3.07 & 3.38 & 3.88 & 3.27 & \textbf{1.8} & \underline{2.47} \\
    \hline
    \end{tabular}
}
\end{subtable}%
\end{minipage}

\caption{Plausibility (OS) results.}
\end{table}
\end{center}
}

\vspace{-24pt}
In the case of multivariate datasets (see Table~\ref{tab:results_plausibility_multivariate}), the results are more conclusive. Multi-SpaCE demonstrates superior performance across evaluation models, achieving the lowest average outlier scores under AE and IF, and second-best results under LOF. This consistency is driven by its ability to produce counterfactuals that are ranked first or second across nearly all datasets, highlighting its robustness in generating plausible instances while simultaneously preserving high performance in other evaluation dimensions. COMTE typically ranks second in plausibility, particularly when plausibility is assessed using Autoencoder and Local Outlier Factor models. These results indicate Multi-SpaCE's ability to generate plausible counterfactuals across various scenarios while maintaining perfect validity, unlike most of the competing approaches.

\begin{center}
\setlength{\tabcolsep}{2pt}
\setlength\extrarowheight{-3pt}
\begin{table}[h]
\resizebox{1\textwidth}{!}{ 
    \centering
    \begin{tabular}{l|cccc|cccc|cccc}
    \hline
    \multirow{2}{*}{Dataset} & \multicolumn{4}{c}{OS(AE)} & \multicolumn{4}{c}{OS(IF)} & \multicolumn{4}{c}{OS(LOF)} \\
     & COMTE & AB-CF & DiscoX & Multi-SpaCE & COMTE & AB-CF & DiscoX & Multi-SpaCE & COMTE & AB-CF & DiscoX & Multi-SpaCE \\
    \hline
    AWR & 0.8 & 0.8 & \textbf{0.65} & \underline{0.66} & 0.5 & 0.55 & \textbf{0.32} & \underline{0.48} & 0.32 & 0.43 & \textbf{0.16} & \underline{0.3} \\
    BasicMotions & 0.51 & 0.71 & \textbf{0.23} & \underline{0.33} & 0.38 & 0.7 & \textbf{0.1} & \underline{0.28} & 0.16 & 0.54 & \textbf{0.0} & \underline{0.05} \\
    Cricket & \textbf{0.71} & 0.78 & - & \textbf{0.71} & \underline{0.31} & 0.43 & - & \textbf{0.29} & \textbf{0.12} & 0.14 & - & \underline{0.13} \\
    Epilepsy & \textbf{0.48} & \underline{0.56} & 0.64 & \underline{0.56} & \textbf{0.25} & 0.29 & 0.31 & \underline{0.28} & \textbf{0.1} & \textbf{0.1} & 0.12 & 0.11 \\
    NATOPS & 0.82 & \underline{0.72} & 0.74 & \textbf{0.61} & 0.3 & 0.36 & \textbf{0.27} & \underline{0.29} & 0.24 & 0.39 & \textbf{0.12} & \underline{0.2} \\
    PEMS-SF & \underline{0.52} & 0.54 & - & \textbf{0.51} & \underline{0.23} & \underline{0.23} & - & \textbf{0.22} & \underline{0.06} & 0.12 & - & \textbf{0.05} \\
    PenDigits & \textbf{0.2} & 0.4 & 0.56 & \underline{0.26} & \textbf{0.29} & 0.35 & 0.41 & \underline{0.32} & \textbf{0.22} & 0.42 & 0.66 & \underline{0.23} \\
    RacketSports & 0.72 & \textbf{0.53} & 0.78 & \underline{0.69} & 0.35 & \textbf{0.2} & 0.37 & \underline{0.33} & \underline{0.12} & \textbf{0.07} & 0.16 & \underline{0.12} \\
    SR-SCP1 & 0.61 & \textbf{0.42} & 0.49 & \underline{0.45} & \underline{0.2} & 0.21 & \textbf{0.19} & 0.22 & \underline{0.11} & \textbf{0.07} & 0.12 & 0.13 \\
    UWave & \textbf{0.49} & 0.72 & 0.75 & \underline{0.61} & \textbf{0.44} & 0.5 & \textbf{0.44} & \textbf{0.44} & \underline{0.37} & 0.46 & \textbf{0.36} & 0.4 \\
    \hline
    Average Rank & \underline{2.3} & 2.5 & 3.0 & \textbf{1.7} & \underline{2.1} & 3.1 & 2.12 & \textbf{1.9} & \textbf{2.0} & 2.8 & 2.38 & \underline{2.3} \\
    \hline
    \end{tabular}
}

\caption{Outlier Score (OS) results in Multivariate datasets. The lower the OS the higher the plausibility.}
\label{tab:results_plausibility_multivariate}
\end{table}
\end{center}

\mycomment{
\begin{center}

\setlength{\tabcolsep}{2pt}
\begin{table}[ht]

\begin{subtable}[t]{.45\textwidth}
\subcaption{Multivariate Datasets}
\resizebox{0.98\textwidth}{!}{ 
    \centering
    \begin{tabular}{l|cccc}
    \multirow{2}{*}{Dataset} & \multicolumn{4}{c}{OS(AE)} \\
     & COMTE & AB-CF & DiscoX & Multi-SpaCE \\
    \hline
    AWR & 0.78 & 0.74 & \textbf{0.67} & \underline{0.7} \\
    BasicMotions & 0.49 & \underline{0.4} & 0.43 & \textbf{0.34} \\
    Cricket & \underline{0.75} & \textbf{0.74} & - & 0.76 \\
    Epilepsy & \textbf{0.48} & 0.59 & \underline{0.58} & 0.61 \\
    NATOPS & 0.79 & \textbf{0.62} & 0.71 & \underline{0.65} \\
    PEMS-SF & 0.54 & \textbf{0.5} & - & \underline{0.51} \\
    PenDigits & \textbf{0.19} & 0.3 & 0.56 & \underline{0.27} \\
    RacketSports & \underline{0.74} & \underline{0.74} & 0.75 & \textbf{0.73} \\
    SR-SCP1 & 0.62 & \textbf{0.42} & 0.61 & \underline{0.44} \\
    UWave & \textbf{0.45} & 0.72 & 0.72 & \underline{0.62} \\
    \hline
    Average Rank & 2.6 & \textbf{2.0} & 2.88 & \underline{2.1} \\
    \hline
    \end{tabular}
}
\end{subtable}%
\begin{subtable}[t]{.55\textwidth}
\subcaption{Univariate Datasets}
\resizebox{0.98\textwidth}{!}{ 
    \centering
    \begin{tabular}{l|cccccc}
    \multirow{2}{*}{Dataset} & \multicolumn{6}{c}{OS(AE)} \\
     & NG & Glacier & Glacier(AE) & AB-CF & DiscoX & Multi-SpaCE \\
    \hline
    Coffee & 0.98 & 0.98 & \textbf{0.32} & 0.98 & \underline{0.96} & 0.98 \\
    ECG200 & \underline{0.69} & 0.84 & \textbf{0.39} & 0.85 & 1.02 & 0.86 \\
    FordA & \underline{0.35} & 0.39 & \textbf{0.1} & 0.38 & 0.4 & 0.47 \\
    Gunpoint & \underline{0.77} & 0.78 & \textbf{0.56} & 0.81 & 0.86 & 0.79 \\
    HandOutlines & \textbf{0.29} & 0.95 & 0.35 & \underline{0.33} & 0.48 & 0.48 \\
    ItalyPower & \underline{0.61} & 0.63 & \textbf{0.27} & 0.69 & 0.97 & 0.65 \\
    PPOC & 0.14 & 0.13 & \textbf{0.03} & \underline{0.08} & 0.52 & 0.16 \\
    Strawberry & 0.35 & \underline{0.28} & \textbf{0.18} & 0.41 & 0.74 & 0.32 \\
    CBF & \underline{0.93} & - & - & 0.99 & \textbf{0.87} & \underline{0.93} \\
    CinCECGTorso & 0.86 & - & - & \underline{0.75} & \textbf{0.6} & 0.77 \\
    TwoPatterns & \underline{0.74} & - & - & 0.75 & 1.03 & \textbf{0.65} \\
    ECG5000 & \underline{0.39} & - & - & 0.54 & 0.55 & \textbf{0.38} \\
    Plane & \textbf{0.5} & - & - & 0.79 & 1.09 & \underline{0.54} \\
    FacesUCR & \textbf{0.53} & - & - & 0.6 & 0.65 & \underline{0.55} \\
    NI-ECG2 & \textbf{0.1} & - & - & \textbf{0.1} & 0.13 & 0.12 \\
    \hline
    Average Rank & \underline{2.2} & 3.38 & \textbf{1.25} & 3.2 & 4.2 & 3.2 \\
    \hline
    \end{tabular}
}
\end{subtable}%
\caption{Plausibility results calculated using the Autoencoder.}
\label{tab:results_plausibility_AE}
\end{table}
\end{center}

\begin{center}

\setlength{\tabcolsep}{2pt}
\begin{table}[ht]

\begin{subtable}[t]{.45\textwidth}
\subcaption{Multivariate Datasets}
\resizebox{0.98\textwidth}{!}{ 
    \centering
    \begin{tabular}{l|cccc}
    \multirow{2}{*}{Dataset} & \multicolumn{4}{c}{OS(IF)} \\
     & COMTE & AB-CF & DiscoX & Multi-SpaCE \\
    \hline
    AWR & \textbf{0.47} & \underline{0.51} & 0.54 & \underline{0.51} \\
    BasicMotions & \textbf{0.61} & \underline{0.64} & 0.73 & 0.72 \\
    Cricket & \textbf{0.6} & \underline{0.65} & - & 0.67 \\
    Epilepsy & 0.76 & \underline{0.69} & 0.7 & \textbf{0.66} \\
    NATOPS & \textbf{0.69} & 0.75 & 0.74 & \underline{0.72} \\
    PEMS-SF & \textbf{0.77} & 0.81 & - & \underline{0.78} \\
    PenDigits & 0.73 & \underline{0.66} & \textbf{0.6} & 0.68 \\
    RacketSports & 0.65 & \textbf{0.62} & \underline{0.64} & \underline{0.64} \\
    SR-SCP1 & 0.81 & \underline{0.8} & 0.81 & \textbf{0.79} \\
    UWave & 0.63 & 0.57 & \underline{0.54} & \textbf{0.52} \\
    \hline
    Average Rank & 2.4 & \underline{2.3} & 2.75 & \textbf{2.0} \\
    \hline
    \end{tabular}
}
\end{subtable}%
\begin{subtable}[t]{.55\textwidth}
\subcaption{Univariate Datasets}
\resizebox{0.98\textwidth}{!}{ 
    \centering
    \begin{tabular}{l|cccccc}
    \multirow{2}{*}{Dataset} & \multicolumn{6}{c}{OS(IF)} \\
     & NG & Glacier & Glacier(AE) & AB-CF & DiscoX & Multi-SpaCE \\
    \hline
    Coffee & 0.68 & \underline{0.58} & \textbf{-0.32} & 0.69 & 0.6 & 0.63 \\
    ECG200 & 0.67 & 0.66 & 0.82 & 0.68 & \textbf{0.6} & \underline{0.63} \\
    FordA & 0.53 & \underline{0.46} & 0.64 & 0.47 & 0.88 & \textbf{0.45} \\
    Gunpoint & 0.8 & 0.81 & \textbf{0.68} & 0.8 & \underline{0.75} & 0.8 \\
    HandOutlines & \underline{0.88} & 0.89 & 0.89 & \underline{0.88} & \textbf{0.83} & \underline{0.88} \\
    ItalyPower & 0.76 & \textbf{0.68} & 0.74 & 0.75 & \underline{0.7} & 0.76 \\
    PPOC & 0.84 & 0.81 & 0.93 & 0.9 & \textbf{0.67} & \underline{0.8} \\
    Strawberry & 0.85 & \underline{0.84} & 0.87 & 0.88 & \textbf{0.79} & 0.85 \\
    CBF & 0.58 & - & - & \textbf{0.49} & 0.55 & \underline{0.54} \\
    CinCECGTorso & \textbf{0.58} & - & - & \underline{0.6} & 0.67 & 0.63 \\
    TwoPatterns & 0.64 & - & - & \underline{0.63} & \textbf{0.5} & 0.65 \\
    ECG5000 & 0.68 & - & - & \textbf{0.57} & \underline{0.64} & 0.74 \\
    Plane & 0.76 & - & - & \underline{0.72} & \textbf{0.54} & 0.75 \\
    FacesUCR & 0.57 & - & - & \textbf{0.53} & \underline{0.54} & 0.55 \\
    NI-ECG2 & 0.82 & - & - & 0.83 & \textbf{0.71} & \underline{0.8} \\
    \hline
    Average Rank & 3.47 & 3.0 & 4.0 & 3.13 & \textbf{2.07} & \underline{2.87} \\
    \hline
    \end{tabular}
}
\end{subtable}%
\caption{Plausibility results calculated using Isolation Forest.}
\label{tab:results_plausibility_IF}
\end{table}
\end{center}

\begin{center}

\setlength{\tabcolsep}{2pt}
\begin{table}[ht]

\begin{subtable}[t]{.45\textwidth}
\subcaption{Multivariate Datasets}
\resizebox{0.98\textwidth}{!}{ 
    \centering
    \begin{tabular}{l|cccc}
    \multirow{2}{*}{Dataset} & \multicolumn{4}{c}{OS(LOF)} \\
     & COMTE & AB-CF & DiscoX & Multi-SpaCE \\
    \hline
    AWR & \textbf{0.65} & \underline{0.68} & 0.73 & 0.7 \\
    BasicMotions & \textbf{0.78} & \underline{0.86} & 0.94 & 0.94 \\
    Cricket & \textbf{0.75} & \underline{0.82} & - & 0.86 \\
    Epilepsy & 0.93 & 0.88 & \underline{0.86} & \textbf{0.84} \\
    NATOPS & \textbf{0.76} & 0.81 & \underline{0.79} & \underline{0.79} \\
    PEMS-SF & \underline{0.94} & \textbf{0.92} & - & 0.95 \\
    PenDigits & 0.78 & \underline{0.68} & \textbf{0.37} & 0.75 \\
    RacketSports & 0.87 & \textbf{0.86} & \textbf{0.86} & \textbf{0.86} \\
    SR-SCP1 & 0.91 & 0.9 & \textbf{0.84} & \underline{0.89} \\
    UWave & 0.61 & \textbf{0.58} & 0.6 & \textbf{0.58} \\
    \hline
    Average Rank & 2.6 & \textbf{2.1} & \underline{2.12} & 2.2 \\
    \hline
    \end{tabular}
}
\end{subtable}%
\begin{subtable}[t]{.55\textwidth}
\subcaption{Univariate Datasets}
\resizebox{0.98\textwidth}{!}{ 
    \centering
    \begin{tabular}{l|cccccc}
    \multirow{2}{*}{Dataset} & \multicolumn{6}{c}{OS(LOF)} \\
     & NG & Glacier & Glacier(AE) & AB-CF & DiscoX & Multi-SpaCE \\
    \hline
    Coffee & 0.71 & 0.62 & \textbf{-25.48} & 0.74 & \underline{-3.86} & 0.6 \\
    ECG200 & 0.74 & 0.8 & 0.82 & \underline{0.53} & \textbf{0.06} & 0.66 \\
    FordA & 0.74 & \underline{0.65} & 0.7 & 0.73 & 0.75 & \textbf{0.6} \\
    Gunpoint & \underline{0.9} & 0.91 & 0.92 & 0.91 & \textbf{0.84} & \underline{0.9} \\
    HandOutlines & \underline{0.97} & 0.98 & 0.98 & \underline{0.97} & \textbf{0.89} & \underline{0.97} \\
    ItalyPower & 0.89 & \textbf{0.81} & 0.89 & 0.9 & \underline{0.86} & 0.89 \\
    PPOC & \underline{0.82} & 0.93 & 0.96 & 0.96 & \textbf{-0.17} & 0.9 \\
    Strawberry & \underline{0.89} & \underline{0.89} & \underline{0.89} & 0.9 & \textbf{0.4} & \underline{0.89} \\
    CBF & \textbf{0.84} & - & - & \underline{0.86} & \underline{0.86} & \underline{0.86} \\
    CinCECGTorso & \textbf{-3.25} & - & - & \underline{-0.92} & -0.09 & -0.29 \\
    TwoPatterns & 0.71 & - & - & 0.69 & \textbf{0.41} & \underline{0.67} \\
    ECG5000 & 0.77 & - & - & \underline{0.66} & \textbf{0.57} & 0.75 \\
    Plane & 0.93 & - & - & \underline{0.87} & \textbf{0.63} & 0.91 \\
    FacesUCR & 0.73 & - & - & \textbf{0.71} & \underline{0.72} & 0.73 \\
    NI-ECG2 & 0.98 & - & - & \underline{0.97} & \textbf{0.87} & \underline{0.97} \\
    \hline
    Average Rank & 3.07 & 3.38 & 3.88 & 3.27 & \textbf{1.8} & \underline{2.47} \\
    \hline
    \end{tabular}
}
\end{subtable}%
\caption{Plausibility results calculated using Local Outlier Factor.}
\label{tab:results_plausibility_LOF}
\end{table}
\end{center}
}

\begin{figure}[H]
    \centering
    \vspace*{-6pt}
    \begin{subfigure}[t]{0.85\textwidth}
        \centering
        \includegraphics[width=1\textwidth]{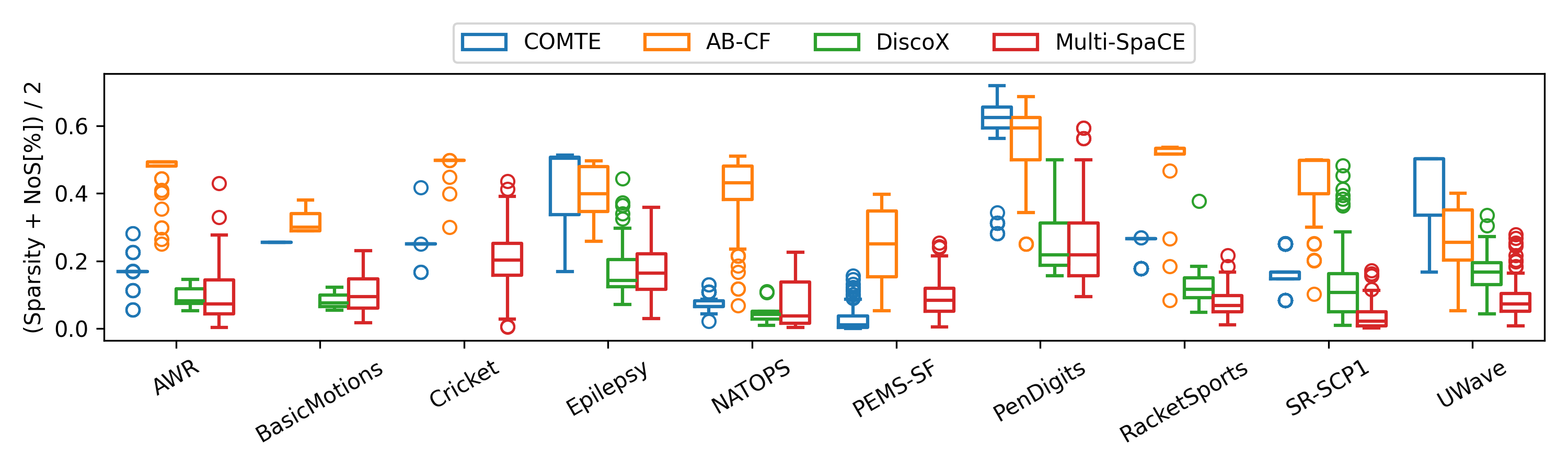}
        \caption{Multivariate Datasets.}
    \end{subfigure}
    
    \begin{subfigure}[t]{1\textwidth}
        \centering
        \includegraphics[width=0.9\textwidth]{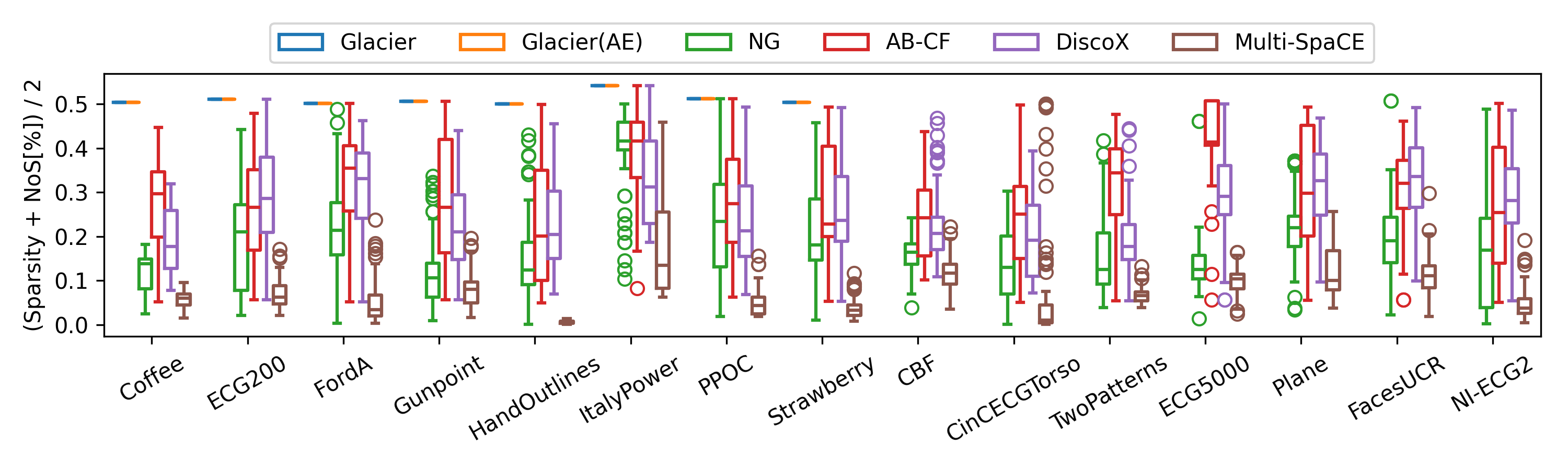}
        \caption{Univariate Datasets.}
    \end{subfigure}
    \caption{Arithmetic mean between normalized sparsity and normalized number of subsequences.}
    \label{fig:sparsity-subsequences}
\end{figure}

Sparsity and contiguity are closely related properties. Methods that restrict solutions to one subsequence per channel, such as NG and COMTE, excel in contiguity. However, these approaches often result in unnecessary changes, reflected in worse sparsity scores (see Supplementary Material). Therefore, sparsity and the number of sub-sequences should be measured jointly. To assess the different methods following this intuition, we simply compute the arithmetic mean between the normalized sparsity and the normalized number of subsequences. Ideally, the metric value should be close to 0, indicating both low sparsity and a minimal number of subsequences. Figure~\ref{fig:sparsity-subsequences} presents the results as boxplots. 
For univariate datasets, Multi-SpaCE achieves the lowest median value, consistently outperforming other methods. In multivariate datasets, Multi-SpaCE achieves the best median value, with a few exceptions like COMTE in \textit{PEMS-SF} and DiscoX in \textit{BasicMotions} and \textit{Epilepsy}. These results demonstrate Multi-SpaCE's ability to balance sparsity and contiguity effectively across diverse datasets.

\begin{figure}[H]
    \centering
    \includegraphics[width=0.9\textwidth]{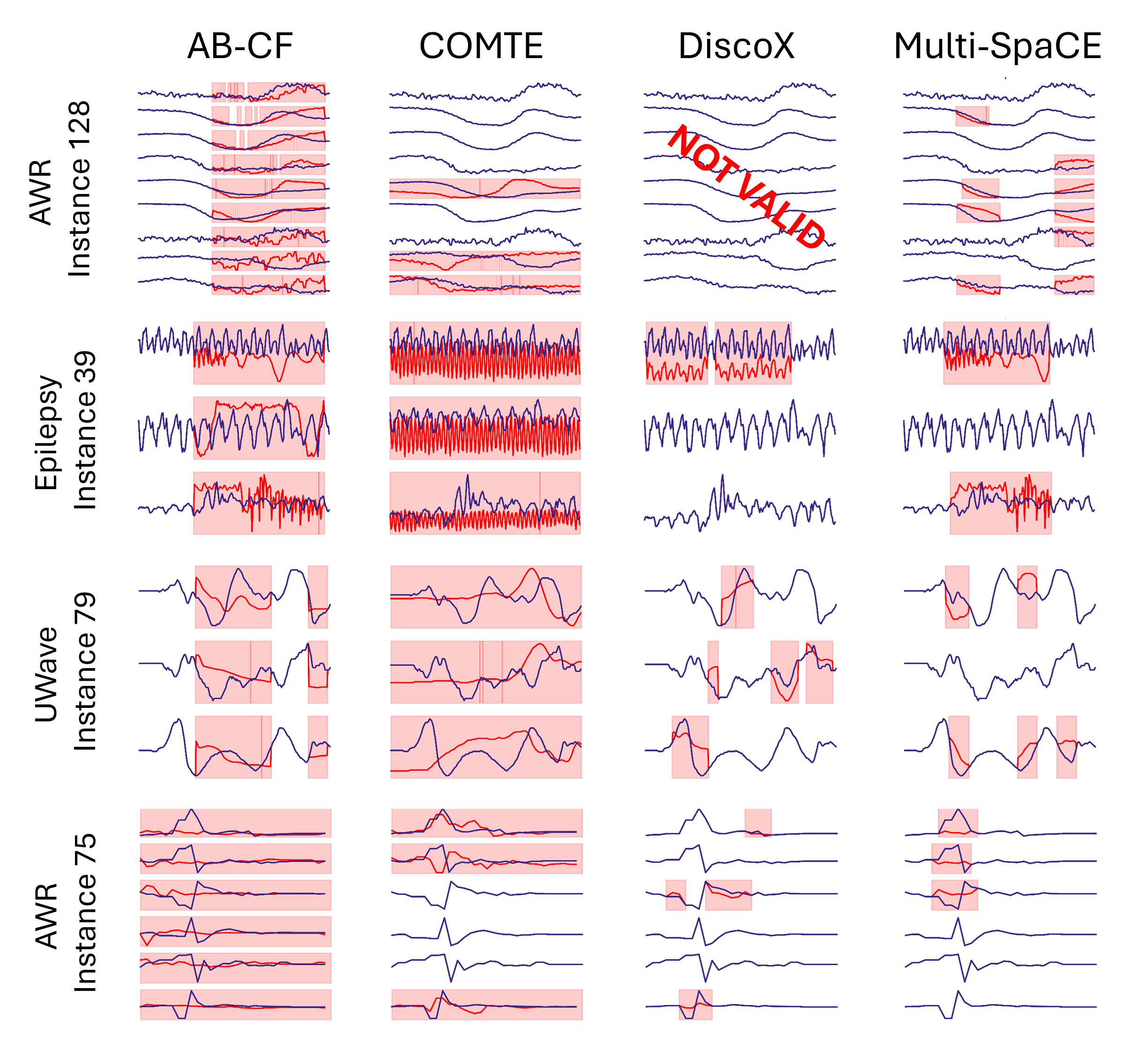}
    \caption{Examples of counterfactual explanations. The red and blue lines represent the original input and the generated counterfactual respectively. The background in red indicates the timestamps in which the counterfactual value differs from the original one. Not valid counterfactuals are likely for some methods (i.e., DiscoX has a validity of 9\% in \textit{AWR}).}
    \label{fig:example_cfs}
\end{figure}

\vspace{-10pt}
Figure~\ref{fig:example_cfs} presents illustrative examples of counterfactual explanations generated by Multi-SpaCE and baseline methods. AB-CF typically produces counterfactuals with a large number of changes but a limited number of subsequences. COMTE, by design, modifies entire channels, while DiscoX tends to generate highly sparse counterfactuals, but with low probability of achieving valid solutions. In contrast, Multi-SpaCE produces counterfactuals that modify a limited number of subsequences, often resulting in smoother and more interpretable transitions compared to the abrupt changes on the original signal observed in methods like DiscoX.

\begin{figure}[H]
    \centering
    \includegraphics[width=0.8\textwidth]{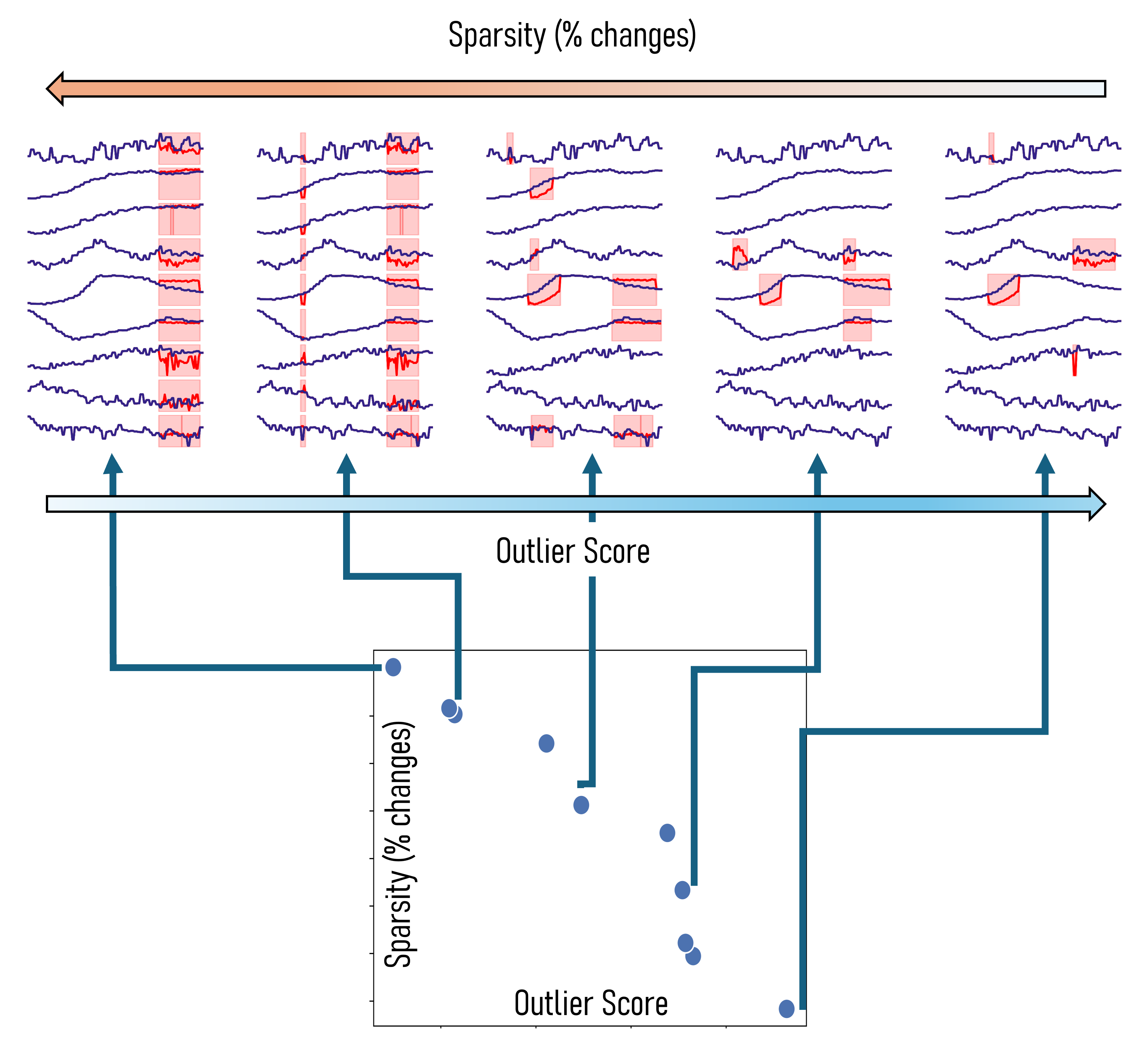}
    \caption{Example of Pareto Front (sparsity vs plausibility) of counterfactual explanations produced by Multi-SpaCE for instance 135 of \textit{ArticularyWordRecognition (AWR)} test set.}
    \label{fig:example_pareto_front}
\end{figure}

It is important to note that the counterfactuals shown in Figure~\ref{fig:example_cfs} correspond to the solutions with the highest utility score, computed using the fixed objective weights described in Section~\ref{sec:setup}. However, recall that Multi-SpaCE returns a Pareto front of non-dominated solutions, allowing the end-user to select the counterfactual that best aligns with their specific interpretability or plausibility requirements. Figure~\ref{fig:example_pareto_front} illustrates an example of such a Pareto front for the sparsity and plausibility objectives. As shown, the most plausible counterfactuals—those with the lowest outlier scores—tend to concentrate changes at the same time steps across all channels. As sparsity is increasingly prioritized, solutions begin to distribute changes more independently across channels, which in turn raises the outlier score and reduces overall plausibility.

\section{Conclusions and Future Work}
\label{sec:conclusions}

This work has introduced Multi-SpaCE, a novel method to generate counterfactual explanations for multivariate time series classification problems. By leveraging multi-objective optimization, Multi-SpaCE balances proximity, sparsity, plausibility, and contiguity, enabling users to tailor explanations to specific preferences.

The experimental results confirm the effectiveness of Multi-SpaCE in diverse datasets, consistently achieving good rankings across all metrics and addressing important limitations of previous approaches, such as the use of fixed-length subsequences and limited multivariate support. Most importantly, Multi-SpaCE ensures perfect validity in univariate and multivariate settings thanks to the inclusion of a strict penalization term, which addresses a critical gap in many contemporary methods. 

Although this work addresses an important gap in time series explainability using counterfactuals, there are several promising directions for future improvements. One area is the use of generative models, such as Variational Autoencoders (VAEs), Generative Adversarial Networks (GANs) or Diffusion Models, to create richer and more realistic modifications of the original instance, rather than relying on substitutions from the NUN. This would improve plausibility and allow for greater diversity in the counterfactuals, particularly in complex datasets. Another promising direction is the inclusion of amortized generation techniques, which could leverage knowledge from previously explained instances to generate counterfactual explanations. This would reduce the computational overhead associated with optimization-based methods, enabling faster and more efficient generation.




\section*{Declarations}
\textbf{Funding} Grant PID2023-153035NB-I00 funded by
MICIU/AEI/10.13039/501100011033 and ERDF/EU.

\textbf{Availability of data and material} The experimental validation was done using public datasets from the UCR and UEA archives \cite{UCRArchive, UEAArchive}. 

\textbf{Code availability} at \url{https://github.com/MarioRefoyo/Multi-SpaCE}

\textbf{Authors' contributions} M.R. performed the experiments, analyzed the results, and wrote the paper. D.L. gave conceptual and technical advice, supervised the analysis, and edited the manuscript.


\bibliography{sn-bibliography}% common bib file

\newpage
\begin{appendices}
\section{Mask and mutation ablation study}\label{app:ablation}
We evaluated the influence of mutation parameters and the different types of masks proposed in Section~\ref{sec:multivariate_mask}, as these components are central to our framework. The comparisons were performed in the order of presentation.

\subsection{Extension/Compression mutation probability}

The first experiment investigated the impact of extension and compression mutation probabilities under the common and independent mask settings. The objective was to identify optimal probabilities and assess the sensitivity of the optimization process to this parameter. The tested probabilities were $p^e=p^c \in \{0.1, 0.25, 0.5, 0.75, 0.9\}$, while $p^p$ was fixed at 0. The NSGA-II parameters were set to a population size of $N=100$ and a total of $G=100$ generations.

In terms of validity, all configurations achieved a perfect score. Regarding proximity, Table~\ref{tab:ablation_proximity_ec} shows that higher values perform best for the common mask setting, whereas higher intermediate values are preferable for the independent mask setting. Regarding plausibility, as shown in Table~\ref{tab:ablation_plausibility_ec}, intermediate probabilities yield better results for both mask types.

\begin{center}
\setlength{\tabcolsep}{6pt}
\setlength\extrarowheight{-3pt}
\begin{table}[ht]
\centering
\begin{subtable}[t]{0.8\textwidth}
\resizebox{1\textwidth}{!}{ 
    \centering
    \begin{tabular}{l|ccccc}
    \hline
    \multirow{2}{*}{Dataset} & \multicolumn{5}{c}{Common mask} \\
     & $p^e=p^c=0.1$ & $p^e=p^c=0.25$ & $p^e=p^c=0.5$ & $p^e=p^c=0.75$ & $p^e=p^c=0.9$ \\
    \hline
    AWR & 19.3 & 19.38 & \underline{19.24} & 19.26 & \textbf{19.2} \\
    BasicMotions & 68.39 & 70.1 & 68.66 & \underline{68.23} & \textbf{68.02} \\
    Cricket & 57.38 & \underline{54.68} & \textbf{53.5} & 55.53 & 58.99 \\
    Epilepsy & 14.06 & 14.13 & 14.22 & \underline{14.03} & \textbf{13.93} \\
    NATOPS & 9.57 & 9.56 & 9.56 & \underline{9.54} & \textbf{9.52} \\
    RacketSports & \textbf{61.55} & 61.69 & \underline{61.62} & 61.71 & 61.65 \\
    SR-SCP1 & \underline{416.63} & 422.63 & 417.54 & \textbf{410.61} & 424.47 \\
    UWave & 17.43 & 17.7 & 17.66 & \underline{17.41} & \textbf{16.78} \\
    \hline
    Average Rank & 3.12 & 4.0 & 3.0 & \underline{2.5} & \textbf{2.25} \\
    \hline
    \end{tabular}
}
\end{subtable}
\vspace*{6pt}

\begin{subtable}[t]{0.8\textwidth}
\resizebox{1\textwidth}{!}{ 
    \centering
    \begin{tabular}{l|ccccc}
    \hline
    \multirow{2}{*}{Dataset} & \multicolumn{5}{c}{Independent mask} \\
     & $p^e=p^c=0.1$ & $p^e=p^c=0.25$ & $p^e=p^c=0.5$ & $p^e=p^c=0.75$ & $p^e=p^c=0.9$ \\
    \hline
    AWR & 19.18 & \underline{17.92} & \textbf{17.86} & 18.03 & 18.65 \\
    BasicMotions & 71.71 & 70.5 & \textbf{68.4} & \underline{69.9} & 70.28 \\
    Cricket & 64.93 & \underline{62.15} & \textbf{61.99} & 63.04 & 66.47 \\
    Epilepsy & 13.49 & \textbf{13.37} & 13.51 & \underline{13.46} & 13.62 \\
    NATOPS & 8.97 & 8.86 & \underline{8.52} & \textbf{8.46} & 8.59 \\
    RacketSports & 51.23 & 50.15 & 49.9 & \underline{49.56} & \textbf{48.87} \\
    SR-SCP1 & 465.32 & 419.69 & \underline{394.73} & \textbf{386.52} & 406.45 \\
    UWave & 16.17 & 16.23 & \textbf{15.68} & \underline{15.95} & 16.15 \\
    \hline
    Average Rank & 4.5 & 3.25 & \textbf{1.88} & \underline{2.0} & 3.38 \\
    \hline
    \end{tabular}
}
\end{subtable}

\caption{Proximity results for Common/Independent mask structures with different extension/compression mutation probabilities.}
\label{tab:ablation_proximity_ec}
\end{table}
\end{center}

\begin{center}
\setlength{\tabcolsep}{6pt}
\setlength\extrarowheight{-3pt}
\begin{table}[H]
\centering
\begin{subtable}[t]{0.8\textwidth}
\resizebox{1\textwidth}{!}{ 
    \centering
    \begin{tabular}{l|ccccc}
    \hline
    \multirow{2}{*}{Dataset} & \multicolumn{5}{c}{Common mask} \\
     & $p^e=p^c=0.1$ & $p^e=p^c=0.25$ & $p^e=p^c=0.5$ & $p^e=p^c=0.75$ & $p^e=p^c=0.9$ \\
    \hline
    AWR & 0.65 & \textbf{0.64} & \textbf{0.64} & 0.65 & 0.65 \\
    BasicMotions & 0.27 & \textbf{0.26} & \textbf{0.26} & \textbf{0.26} & 0.27 \\
    Cricket & 0.73 & 0.69 & \textbf{0.67} & \underline{0.68} & 0.8 \\
    Epilepsy & 0.56 & \underline{0.54} & \textbf{0.53} & \underline{0.54} & 0.56 \\
    NATOPS & \textbf{0.57} & \textbf{0.57} & \textbf{0.57} & \textbf{0.57} & \textbf{0.57} \\
    RacketSports & \underline{0.62} & \underline{0.62} & \underline{0.62} & \textbf{0.61} & \underline{0.62} \\
    SR-SCP1 & \textbf{0.43} & \textbf{0.43} & \textbf{0.43} & \textbf{0.43} & 0.44 \\
    UWave & 0.62 & \underline{0.6} & \textbf{0.59} & \underline{0.6} & 0.63 \\
    \hline
    Average Rank & 2.88 & \underline{1.62} & \textbf{1.12} & \underline{1.62} & 3.62 \\
    \hline
    \end{tabular}
}
\end{subtable}
\vspace*{6pt}

\begin{subtable}[t]{0.8\textwidth}
\resizebox{1\textwidth}{!}{ 
    \centering
    \begin{tabular}{l|ccccc}
    \hline
    \multirow{2}{*}{Dataset} & \multicolumn{5}{c}{Independent mask} \\
     & $p^e=p^c=0.1$ & $p^e=p^c=0.25$ & $p^e=p^c=0.5$ & $p^e=p^c=0.75$ & $p^e=p^c=0.9$ \\
    \hline
    AWR & 0.75 & \underline{0.71} & \textbf{0.7} & \underline{0.71} & 0.77 \\
    BasicMotions & \textbf{0.32} & \underline{0.33} & \underline{0.33} & \underline{0.33} & \underline{0.33} \\
    Cricket & 0.94 & \underline{0.85} & \textbf{0.84} & 0.93 & 1.05 \\
    Epilepsy & 0.59 & \underline{0.57} & \textbf{0.56} & \underline{0.57} & 0.59 \\
    NATOPS & 0.71 & \underline{0.68} & 0.69 & \textbf{0.67} & 0.69 \\
    RacketSports & \textbf{0.7} & \underline{0.71} & \underline{0.71} & \underline{0.71} & 0.72 \\
    SR-SCP1 & 0.52 & \textbf{0.51} & \textbf{0.51} & 0.52 & 0.53 \\
    UWave & 0.77 & 0.65 & \textbf{0.63} & \textbf{0.63} & 0.77 \\
    \hline
    Average Rank & 3.25 & \underline{2.0} & \textbf{1.5} & \underline{2.0} & 4.12 \\
    \hline
    \end{tabular}
}
\end{subtable}

\caption{Plausibility results for Common/Independent mask structures with different extension/compression mutation probabilities.}
\label{tab:ablation_plausibility_ec}
\end{table}
\end{center}

\begin{figure}[H]
    \centering
    \begin{subfigure}[t]{0.8\textwidth}
        \centering
        \includegraphics[width=\textwidth]{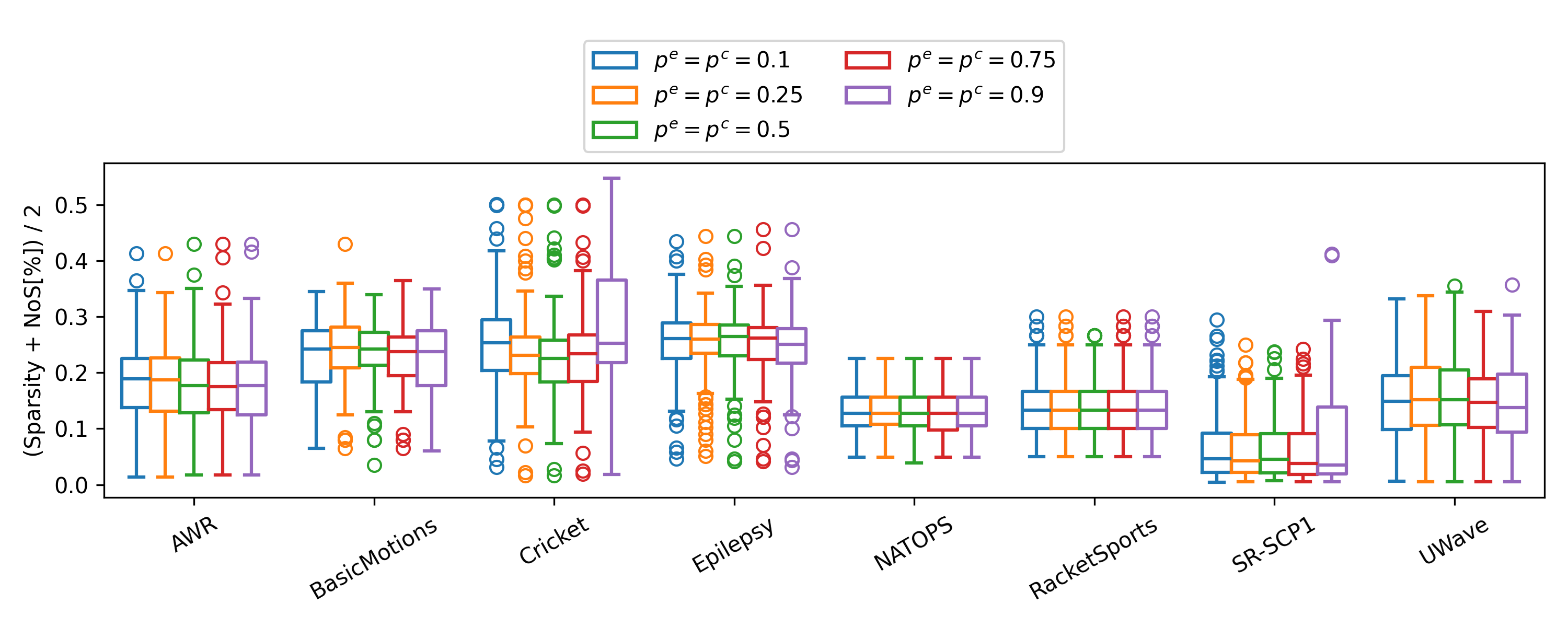}
        \caption{Common mask.}
    \end{subfigure}
    
    \begin{subfigure}[t]{1\textwidth}
        \centering
        \includegraphics[width=0.8\textwidth]{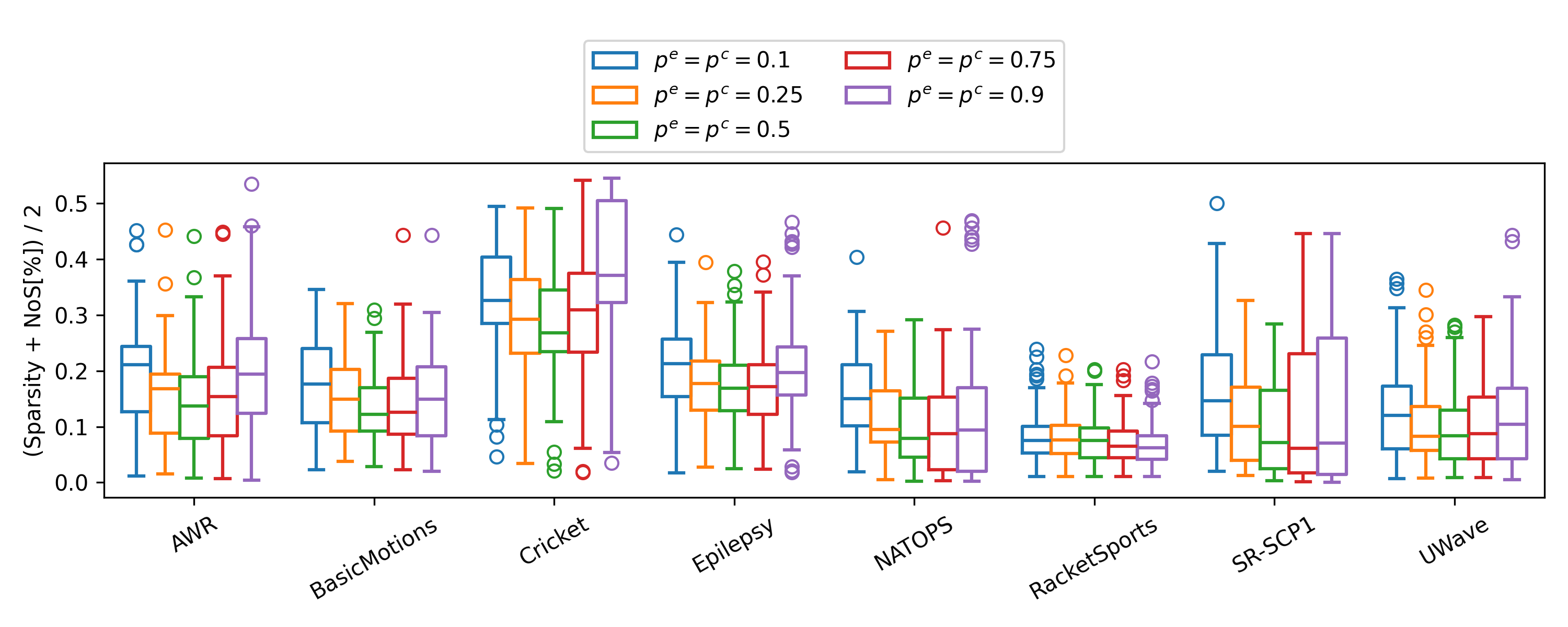}
        \caption{Independent Mask.}
    \end{subfigure}
    \caption{Arithmetic mean between normalized sparsity and normalized number of subsequences depending on the extension/compression mutation probabilities.}
    \label{fig:sparsity-subsequences_ec}
\end{figure}

This behavior is also reflected in the Sparsity-NoS average metric, where intermediate probabilities provide the best balance for both mask settings, as depicted in Figure~\ref{fig:sparsity-subsequences_ec}.

Based on these results, a mutation probability of 0.75 was selected as optimal for the common mask setting, while a probability of 0.5 was chosen for the independent mask setting. These parameters were used in subsequent experiments.

\subsection{Pruning mutation probability}
In this experiment, we introduced a pruning mutation into the configurations identified in the previous section, $p^e=p^c=0.75$ for the common mask, and $p^e=p^c=0.5$ for the independent mask. Specifically, we tested the pruning mutation probabilities $p^p \in \{0.05, 0.1, 0.2, 0.35, 0.5\}$ and also evaluated the case without pruning mutation. The conclusion was clear: not using the pruning mutation yields significantly better results for both mask settings. Considering these findings and the additional computational cost introduced by pruning mutation, we decided to exclude pruning mutation during the same stage as extension/compression mutation.

\subsection{Final iterations using only pruning mutation}
Although the pruning mutation was excluded during the earlier stages of the optimization, we hypothesized that applying it during a final optimization stage could improve contiguity and eliminate unnecessary subsequences. To test this, we divided the optimization process into two stages, as implemented in the final version of Multi-SpaCE shown in Algorithm~\ref{algo:multisubspace}.

The first stage consisted of $G_1=75$ generations with the best parameters identified earlier: $p^e=p^c=0.75$ and $p^p=0$ for the common mask setting, and $p^e=p^c=0.5$ and $p^p=0$ for the independent mask setting. In the second stage, the resulting Pareto front was further optimized over 25 generations ($G_2=25$) using $p^e=p^c=0$ and pruning probabilities $p^p \in \{0.25, 0.5, 0.75\}$. During this stage, the independent mask setting was always used, regardless of the mask setting in the first stage. 

In terms of validity, all configurations achieved perfect scores. Regarding plausibility, the best results were consistently achieved with the common mask with final pruning $p^p=0.75$. For proximity, the common mask with final pruning $p^p=0.5$ performed slightly better than the $p^p=0.75$ setting. For the Sparsity-NoS average, common mask with both final pruning $p^p=0.5$ and $p^p=0.75$ achieved comparable results. Because of the difference in terms plausibility, $p^p=0.75$ is the version used for the experiments in Section~\ref{sec:experiments}.
\begin{center}
\setlength{\tabcolsep}{6pt}
\setlength\extrarowheight{-3pt}
\begin{table}[ht]
\centering
\begin{subtable}[t]{0.8\textwidth}
\resizebox{1\textwidth}{!}{ 
    \centering
    \begin{tabular}{l|ccc|ccc}
    \hline
    \multirow{2}{*}{Dataset} & \multicolumn{3}{c}{Common ($p^e=p^c=0.75$)} & \multicolumn{3}{c}{Independent ($p^e=p^c=0.5$)}\\
    & $p^p=0.25$ & $p^p=0.5$ & $p^p=0.75$ & $p^p=0.25$ & $p^p=0.5$ & $p^p=0.75$ \\
    \hline
    AWR & 17.16 & \textbf{16.33} & 16.52 & 17.09 & \underline{16.51} & 17.12 \\
    BasicMotions & 67.95 & \underline{67.34} & \textbf{66.71} & 68.71 & 68.27 & 68.3 \\
    Cricket & \textbf{48.64} & \underline{48.85} & 51.67 & 54.6 & 59.62 & 60.81 \\
    Epilepsy & \underline{13.01} & \textbf{12.96} & 13.04 & 13.2 & 13.08 & 13.21 \\
    NATOPS & 8.7 & 8.46 & 8.38 & 8.2 & \textbf{7.98} & \underline{8.12} \\
    RacketSports & 50.47 & 50.13 & 50.97 & \underline{48.84} & 48.9 & \textbf{48.69} \\
    SR-SCP1 & 330.68 & \underline{288.91} & \textbf{285.74} & 362.82 & 365.2 & 368.46 \\
    UWave & \underline{15.19} & 15.29 & 15.2 & 15.57 & 15.36 & \textbf{15.1} \\
    \hline
    Average Rank & 3.5 & \textbf{2.62} & \underline{3.0} & 4.25 & 3.62 & 4.0 \\
    \hline
    \end{tabular}
}
\end{subtable}

\caption{Proximity results for Common/Independent mask structures with different pruning mutation probabilities during the final 25 iterations.}
\label{tab:ablation_proximity_fpr}
\end{table}
\end{center}

\begin{center}
\setlength{\tabcolsep}{6pt}
\setlength\extrarowheight{-3pt}
\begin{table}[H]
\centering
\begin{subtable}[t]{0.8\textwidth}
\resizebox{1\textwidth}{!}{ 
    \centering
    \begin{tabular}{l|ccc|ccc}
    \hline
    \multirow{2}{*}{Dataset} & \multicolumn{3}{c}{Common ($p^e=p^c=0.75$)} & \multicolumn{3}{c}{Independent ($p^e=p^c=0.5$)}\\
    & $p^p=0.25$ & $p^p=0.5$ & $p^p=0.75$ & $p^p=0.25$ & $p^p=0.5$ & $p^p=0.75$ \\
    \hline
    AWR & \textbf{0.66} & \textbf{0.66} & \textbf{0.66} & 0.7 & 0.7 & 0.71 \\
    BasicMotions & \underline{0.34} & \underline{0.34} & \textbf{0.33} & 0.35 & 0.35 & 0.35 \\
    Cricket & \underline{0.72} & 0.73 & \textbf{0.71} & 0.85 & 0.88 & 0.88 \\
    Epilepsy & 0.58 & 0.58 & \textbf{0.56} & 0.57 & \textbf{0.56} & 0.57 \\
    NATOPS & \textbf{0.61} & 0.62 & \textbf{0.61} & 0.71 & 0.71 & 0.71 \\
    RacketSports & \underline{0.7} & \underline{0.7} & \textbf{0.69} & 0.71 & 0.71 & 0.71 \\
    SR-SCP1 & \underline{0.46} & \underline{0.46} & \textbf{0.45} & 0.51 & 0.51 & 0.51 \\
    UWave & \textbf{0.59} & \textbf{0.59} & 0.61 & 0.61 & 0.62 & 0.63 \\
    \hline
    Average Rank & \underline{2.0} & 2.38 & \textbf{1.25} & 3.75 & 3.88 & 4.5 \\
    \hline
    \end{tabular}
}
\end{subtable}

\caption{Plausibility results for Common/Independent mask structures with different pruning mutation probabilities during the final 25 iterations.}
\label{tab:ablation_plausibility_fpr}
\end{table}
\end{center}

\begin{figure}[H]
    \centering
    \includegraphics[width=0.8\textwidth]{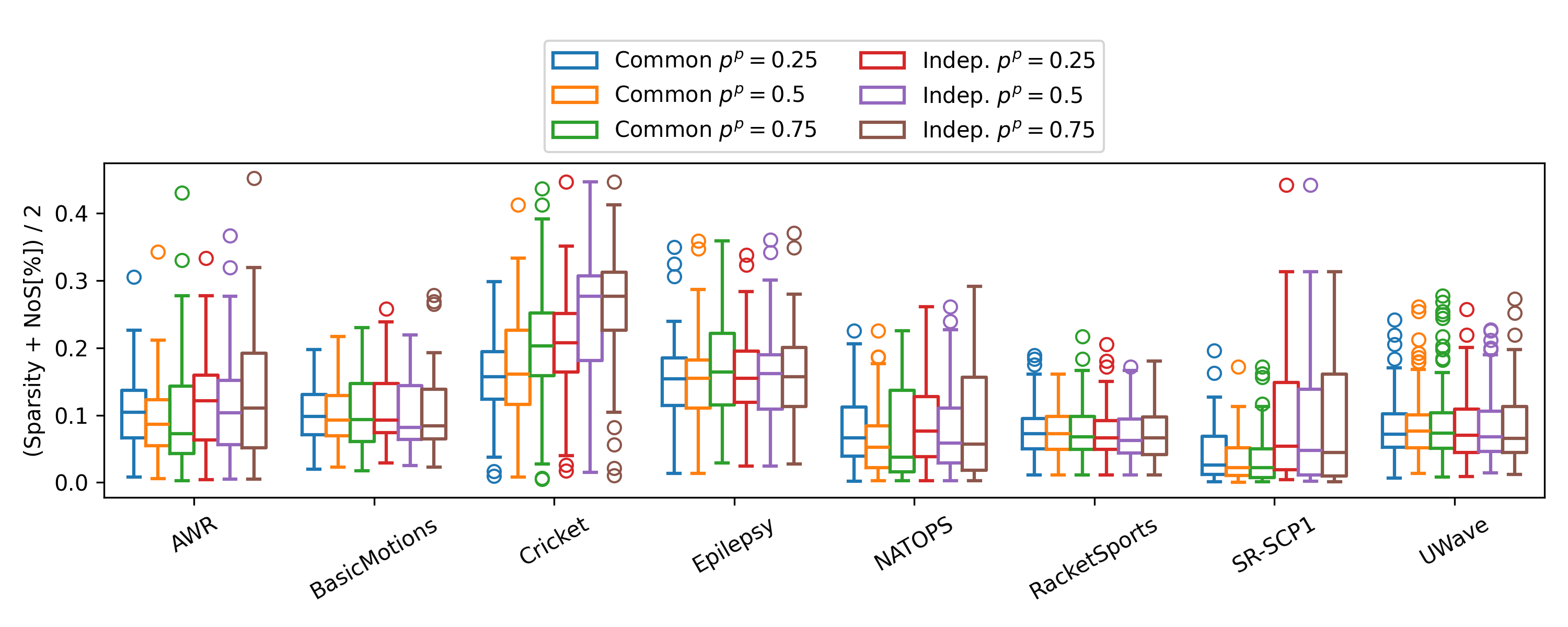}   
    \caption{Arithmetic mean between normalized sparsity and normalized number of subsequences using different pruning mutation probabilities during the final 25 iterations.}
    \label{fig:sparsity-subsequences_fpr}
\end{figure}

\section{Model details}\label{app:model_details}

\subsection{Black-box Classifier}
We trained InceptionTime with default hyperparameters following the implementation in \url{https://github.com/hfawaz/InceptionTime}

\subsection{Autoencoder}
We trained eight different autoencoder architectures for each dataset to better adapt the models to the specific characteristics of the data, serving as a lighter alternative to grid search or Bayesian optimization. Each model randomly splits 10\% of the training data for validation and is trained over 200 epochs, using early stopping with a patience of 30 epochs. Additionally, learning rate reduction is applied when the validation loss plateaus, with a patience of 10 epochs. The initial learning rate is set to 0.001, the batch size to 32, and a dropout rate of 20\% is applied across all models.

Both the encoder and decoder are based on convolutional layers with fixed strides of 2 and include a Dense layer in between, enforcing a target compression rate of the input space to either 6.25\% or 12.5\%. If the dimensionality at the output of the encoder prevents achieving the desired compression rate, the model is not trained, and its performance is represented with a "-" in the results tables. The architectures vary based on the following parameters:

\begin{itemize}
\item \textbf{Shallow:} One convolutional layer with 16 channels and a kernel size of 7. The decoder performs the inverse operations.
\item \textbf{Simple:} Two convolutional layers with 16 and 32 channels, with kernel sizes of 7 and 5, respectively. The decoder performs the inverse operations.
\item \textbf{Intermediate:} Three convolutional layers with 16, 32, and 64 channels, with kernel sizes of 7, 5, and 3, respectively. The decoder performs the inverse operations.
\item \textbf{Complex:} Four convolutional layers with 16, 32, 64, and 128 channels, with kernel sizes of 7, 5, 5, and 3, respectively. The decoder performs the inverse operations.
\end{itemize}

Tables~\ref{tab:ae_multi} and \ref{tab:ae_uni} present the results for each architecture on the multivariate and univariate datasets, respectively.

\begin{center}
\tiny
\begin{table}[H]
\centering
\begin{subtable}[t]{1\textwidth}
\resizebox{0.98\textwidth}{!}{
\begin{tabular}{l|cccccccc}
    \toprule
    Dataset & \thead{Shallow\\(6,25\%)} & \thead{Shallow\\(12,5\%)} & \thead{Simple\\(6,25\%)} & \thead{Simple\\(12,5\%)} & \thead{Interm.\\(6,25\%)} & \thead{Interm.\\(12,5\%)} & \thead{Complex.\\(6,25\%)} & \thead{Complex.\\(12,5\%)} \\
    \midrule
    PEMS-SF & \textbf{0.02} & 0.02 & 0.02 & \underline{0.02} & 0.02 & 0.02 & 0.02 & 0.02 \\
    UWave & - & - & - & 0.38 & 0.39 & \textbf{0.07} & 0.11 & \underline{0.09} \\
    NATOPS & 0.14 & \textbf{0.1} & 0.11 & \underline{0.11} & 0.13 & 0.13 & 0.14 & 0.14 \\
    Cricket & - & 0.56 & 0.49 & 0.32 & 0.3 & 0.21 & \underline{0.19} & \textbf{0.17} \\
    AWR & 0.56 & 0.53 & 0.43 & 0.33 & 0.33 & \textbf{0.23} & 0.28 & \underline{0.28} \\
    Epilepsy & - & - & - & 0.42 & 0.43 & \textbf{0.31} & 0.36 & \underline{0.34} \\
    BasicMotions & - & \underline{1.42} & 1.59 & \textbf{1.34} & 1.63 & 1.58 & 1.91 & 1.79 \\
    RacketSports & - & 2.76 & 2.77 & \textbf{2.57} & 2.76 & \underline{2.73} & 2.81 & 2.76 \\
    PenDigits & - & - & - & 5.03 & 6.57 & \underline{4.25} & 6.29 & \textbf{4.04} \\
    SR-SCP1 & - & 9.93 & \textbf{4.96} & \underline{5.16} & 6.06 & 7.48 & 6.05 & 5.94 \\
    \bottomrule
\end{tabular}
}
\end{subtable}
\caption{Autoencoder reconstruction errors on multivariate test sets.}
\label{tab:ae_multi}
\end{table}

\end{center}

\begin{center}
\small
\begin{table}[H]
\centering
\begin{tabular}{l|ccc}
    \toprule
    Dataset & \thead{Interm.\\(12,5\%)} & \thead{Complex\\(6,25\%)} & \thead{Complex\\(12,5\%)} \\
    \midrule
    HandOutlines & \textbf{0.0} & \underline{0.0} & 0.0 \\
    Strawberry & \underline{0.02} & 0.02 & \textbf{0.01} \\
    NonInvasiveFatalECGThorax2 & \underline{0.02} & 0.03 & \textbf{0.02} \\
    ProximalPhalanxOutlineCorrect & \underline{0.02} & 0.02 & \textbf{0.02} \\
    Plane & \underline{0.09} & 0.1 & \textbf{0.08} \\
    ECG5000 & \underline{0.08} & 0.1 & \textbf{0.08} \\
    CinCECGTorso & \textbf{0.11} & 0.16 & \underline{0.15} \\
    FordA & \textbf{0.16} & 0.27 & \underline{0.16} \\
    ECG200 & \textbf{0.17} & 0.2 & \underline{0.18} \\
    ItalyPowerDemand & 0.26 & \textbf{0.19} & \underline{0.21} \\
    TwoPatterns & \textbf{0.21} & 0.29 & \underline{0.22} \\
    Gunpoint & 0.93 & \underline{0.29} & \textbf{0.28} \\
    Phoneme & \textbf{0.28} & 0.44 & \underline{0.3} \\
    FacesUCR & \underline{0.44} & 0.52 & \textbf{0.41} \\
    Coffee & \textbf{0.63} & \underline{0.73} & 0.82 \\
    CBF & \textbf{0.68} & 0.75 & \underline{0.75} \\
    \bottomrule
\end{tabular}
\caption{Autoencoder reconstruction errors on univariate test sets.}
\label{tab:ae_uni}
\end{table}

\end{center}

\newpage
\subsection{Isolation Forest}
We conduct a grid search over the following parameters to optimize the Isolation Forest (IF) models:
\begin{itemize}
\item n\_estimators: [100, 200, 400, 500]
\item contamination: [0.05, 0.1, 0.2, 0.4]
\item max\_features: [0.1, 0.2, 0.4, 0.5]
\end{itemize}

The model with the highest silhouette score on the test set is selected. The chosen model is then used to calculate the Outlier Score on the main paper. This process is performed independently for each dataset.

\subsection{Local Outlier Factor}
We conduct a grid search over the following parameters to optimize the Local Outlier Factor (LOF) models:
\begin{itemize}
    \item n\_neighbors: [1, 5, 10, 20, 50].
    \item contamination: [0.05, 0.1, 0.2, 0.4].
    \item p: [1, 2].
\end{itemize}

The model with the highest silhouette score on the test set is selected. The chosen model is then used to calculate the Outlier Score, on the main paper. This process is performed independently for each dataset.

\newpage
\section{Sparsity and Contiguity results}
\label{app:sparsity_contiguity}

As discussed on the main paper, sparsity and contiguity must be jointly evaluated. Sparsity minimizes the number of changes applied to the original instance, while contiguity ensures that these changes are coherent. Several methods impose constraints on the structure of changes. For example, NG and COMTE restrict modifications to a single subsequence per channel, resulting in counterfactuals with a low number of subsequences. This is evident in Table~\ref{tab:results_nos}, where these methods achieve the lowest Number of Subsequences scores. Glacier also produces counterfactuals with a single subsequence in the univariate setting, although this outcome is due to the change of every point in the original instance, leading to the worst sparsity scores overall. In the multivariate setting, Multi-SpaCE achieves the second lowest rank. Regarding the univariate setting, it achieves the fourth lowest rank, ranking after NG Glacier(AE) and Glacier methods which use a single subsequence of changes.
\begin{center}
\setlength\extrarowheight{-3pt}
\setlength{\tabcolsep}{2pt}
\begin{table}[ht]

\begin{subtable}[t]{.45\textwidth}
\subcaption{Number of Subsequences (NoS) for Multivariate Datasets}
\resizebox{0.98\textwidth}{!}{ 
    \centering
    \begin{tabular}{l|cccc}
    \hline
    Dataset & COMTE & AB-CF & DiscoX & Multi-SpaCE \\
    \hline
    AWR & \textbf{2.97} & 10.6 & 52.78 & \underline{7.03} \\
    BasicMotions & \textbf{3.0} & 24.11 & 15.33 & \underline{4.4} \\
    Cricket & \textbf{3.0} & \underline{6.88} & - & 57.62 \\
    Epilepsy & \textbf{4.47} & 7.87 & 21.95 & \underline{5.56} \\
    NATOPS & \textbf{3.24} & 47.28 & 33.18 & \underline{10.44} \\
    PEMS-SF & \textbf{86.81} & 1241.13 & - & \underline{1230.05} \\
    PenDigits & 2.87 & 3.13 & \textbf{1.31} & \underline{1.68} \\
    RacketSports & \textbf{2.78} & 8.23 & 10.88 & \underline{2.97} \\
    SR-SCP1 & \textbf{3.93} & \underline{10.69} & 354.09 & 19.39 \\
    UWave & \textbf{2.58} & 6.01 & 30.5 & \underline{3.69} \\
    \hline
    Average Rank & \textbf{1.2} & 3.1 & 3.38 & \underline{2.2} \\
    \hline
    \end{tabular}
}
\end{subtable}%
\begin{subtable}[t]{.55\textwidth}
\subcaption{NoS for Univariate Datasets}
\resizebox{0.98\textwidth}{!}{ 
    \centering
    \begin{tabular}{l|cccccc}
    \hline
    Dataset & NG & Glacier & Glacier(AE) & AB-CF & DiscoX & Multi-SpaCE \\
    \hline
    Coffee & \textbf{1.0} & \textbf{1.0} & \textbf{1.0} & 1.82 & 10.93 & 2.96 \\
    ECG200 & \textbf{1.0} & \textbf{1.0} & \textbf{1.0} & 1.99 & 5.71 & 1.39 \\
    FordA & \textbf{1.0} & \textbf{1.0} & \textbf{1.0} & 2.39 & 10.85 & 4.15 \\
    Gunpoint & \textbf{1.0} & \textbf{1.0} & \textbf{1.0} & 1.91 & 4.95 & 1.48 \\
    HandOutlines & \textbf{1.0} & \textbf{1.0} & \textbf{1.0} & 2.37 & 90.98 & 2.48 \\
    ItalyPower & \textbf{1.0} & \textbf{1.0} & \textbf{1.0} & 2.67 & 1.79 & 1.22 \\
    PPOC & \textbf{1.0} & \textbf{1.0} & \textbf{1.0} & 1.78 & 5.26 & 1.35 \\
    Strawberry & \textbf{1.0} & \textbf{1.0} & \textbf{1.0} & 1.82 & 10.81 & 1.86 \\
    CBF & \textbf{1.0} & - & - & 2.24 & 5.75 & \underline{1.75} \\
    CinCECGTorso & \textbf{1.0} & - & - & \underline{2.26} & 25.7 & 3.9 \\
    TwoPatterns & \textbf{1.0} & - & - & 2.15 & 5.12 & \underline{1.4} \\
    ECG5000 & \textbf{1.0} & - & - & 1.57 & 5.41 & \underline{1.2} \\
    Plane & \textbf{1.0} & - & - & \underline{1.8} & 5.74 & 1.9 \\
    FacesUCR & \textbf{1.0} & - & - & 2.9 & 3.96 & \underline{2.21} \\
    NI-ECG2 & \textbf{1.0} & - & - & \underline{1.99} & 13.39 & 3.5 \\
    \hline
    Average Rank & \textbf{1.0} & \textbf{1.0} & \textbf{1.0} & 3.67 & 5.0 & 3.53 \\
    \hline
    \end{tabular}
}
\end{subtable}%
\caption{NoS results.}
\label{tab:results_nos}
\end{table}
\end{center}

Imposing restrictions on how to change the original instance to form the counterfactual can result in requiring more alterations to achieve the desired outcome, negatively affecting sparsity. Consequently, NG and COMTE perform worse than Multi-SpaCE in terms of sparsity, as shown in Table~\ref{tab:results_sparsity}. Multi-SpaCE consistently achieves the best sparsity performance across univariate datasets and share the lowest average rank with DiscoX for multivariate datasets.
\begin{center}
\setlength\extrarowheight{-3pt}
\setlength{\tabcolsep}{2pt}
\begin{table}[ht]

\begin{subtable}[t]{.45\textwidth}
\subcaption{Multivariate Datasets}
\resizebox{0.98\textwidth}{!}{ 
    \centering
    \begin{tabular}{l|cccc}
    \hline
    Dataset & COMTE & AB-CF & DiscoX & Multi-SpaCE \\
    \hline
    AWR & 0.33 & 0.9 & \textbf{0.11} & \underline{0.19} \\
    BasicMotions & 0.5 & 0.56 & \textbf{0.12} & \underline{0.2} \\
    Cricket & \underline{0.5} & 0.97 & - & \textbf{0.39} \\
    Epilepsy & 0.8 & 0.77 & \textbf{0.27} & \underline{0.32} \\
    NATOPS & 0.14 & 0.73 & \textbf{0.04} & \underline{0.12} \\
    PEMS-SF & \textbf{0.06} & 0.49 & - & \underline{0.15} \\
    PenDigits & 0.82 & 0.72 & \underline{0.35} & \textbf{0.29} \\
    RacketSports & 0.46 & 0.93 & \underline{0.13} & \textbf{0.12} \\
    SR-SCP1 & 0.33 & 0.87 & \underline{0.14} & \textbf{0.06} \\
    UWave & 0.86 & 0.53 & \underline{0.27} & \textbf{0.17} \\
    \hline
    Average Rank & 3.0 & 3.5 & \textbf{1.5} & \textbf{1.5} \\
    \hline
    \end{tabular}
}
\end{subtable}%
\begin{subtable}[t]{.55\textwidth}
\subcaption{Univariate Datasets}
\resizebox{0.98\textwidth}{!}{ 
    \centering
    \begin{tabular}{l|cccccc}
    \hline
    Dataset & NG & Glacier & Glacier(AE) & AB-CF & DiscoX & Multi-SpaCE \\
    \hline
    Coffee & \underline{0.22} & 1.0 & 1.0 & 0.51 & 0.3 & \textbf{0.09} \\
    ECG200 & \underline{0.34} & 1.0 & 1.0 & 0.48 & 0.46 & \textbf{0.11} \\
    FordA & \underline{0.43} & 1.0 & 1.0 & 0.64 & 0.58 & \textbf{0.09} \\
    Gunpoint & \underline{0.23} & 1.0 & 1.0 & 0.56 & 0.37 & \textbf{0.14} \\
    HandOutlines & \underline{0.31} & 1.0 & 1.0 & 0.48 & 0.39 & \textbf{0.01} \\
    ItalyPower & 0.71 & 1.0 & 1.0 & 0.55 & \underline{0.51} & \textbf{0.26} \\
    PPOC & 0.45 & 1.0 & 1.0 & 0.53 & \underline{0.35} & \textbf{0.06} \\
    Strawberry & \underline{0.41} & 1.0 & 1.0 & 0.55 & 0.42 & \textbf{0.06} \\
    CBF & \underline{0.31} & - & - & 0.44 & 0.36 & \textbf{0.2} \\
    CinCECGTorso & \underline{0.27} & - & - & 0.5 & 0.37 & \textbf{0.19} \\
    TwoPatterns & \underline{0.29} & - & - & 0.59 & 0.3 & \textbf{0.11} \\
    ECG5000 & \underline{0.26} & - & - & 0.81 & 0.52 & \textbf{0.19} \\
    Plane & \underline{0.42} & - & - & 0.57 & 0.55 & \textbf{0.21} \\
    FacesUCR & \underline{0.37} & - & - & 0.57 & 0.61 & \textbf{0.19} \\
    NI-ECG2 & \underline{0.33} & - & - & 0.52 & 0.54 & \textbf{0.09} \\
    \hline
    Average Rank & \underline{2.2} & 5.0 & 5.0 & 3.8 & 3.0 & \textbf{1.0} \\
    \hline
    \end{tabular}
}
\end{subtable}%
\caption{Sparsity for Multivariate Datasets.}
\label{tab:results_sparsity}
\end{table}
\end{center}

This relation between contiguity and sparsity motivated the need for their joint evaluation. Multi-SpaCE demonstrates superiority over other methods in the literature when both metrics are considered together.

\newpage
\section{Execution time}\label{app:time}
\begin{figure}[h]
    \centering
    \begin{subfigure}[t]{0.9\textwidth}
        \centering
        \includegraphics[width=1\textwidth]{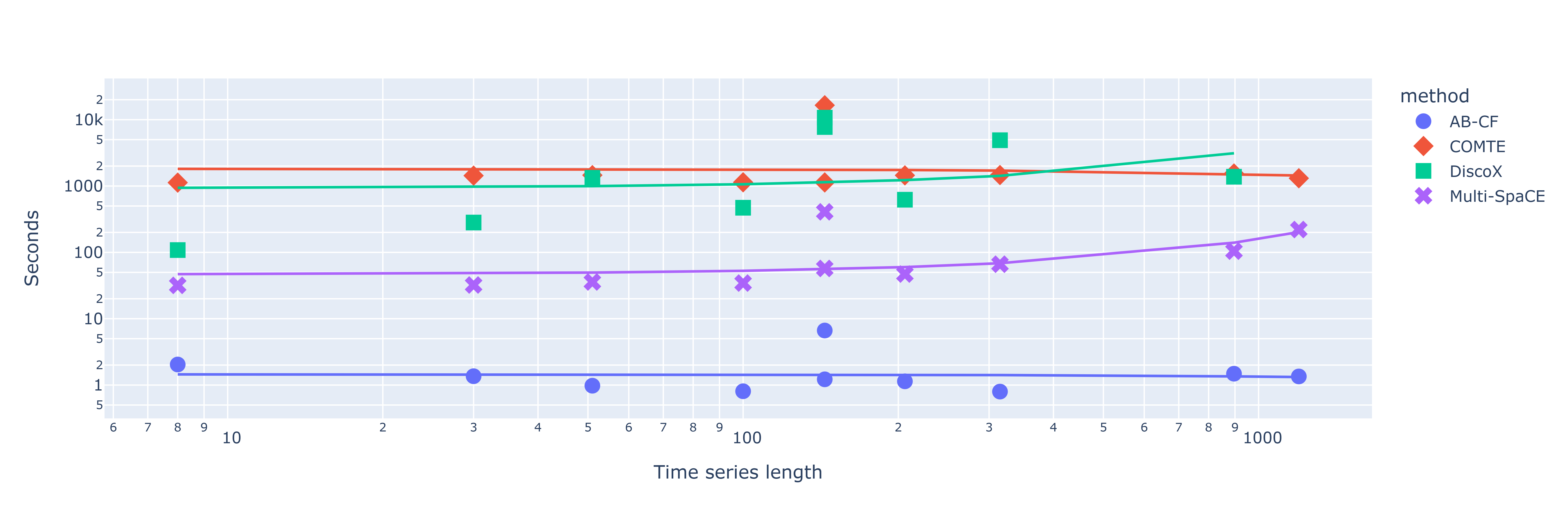}
    \end{subfigure}
    
    \begin{subfigure}[t]{0.9\textwidth}
        \centering
        \includegraphics[width=1\textwidth]{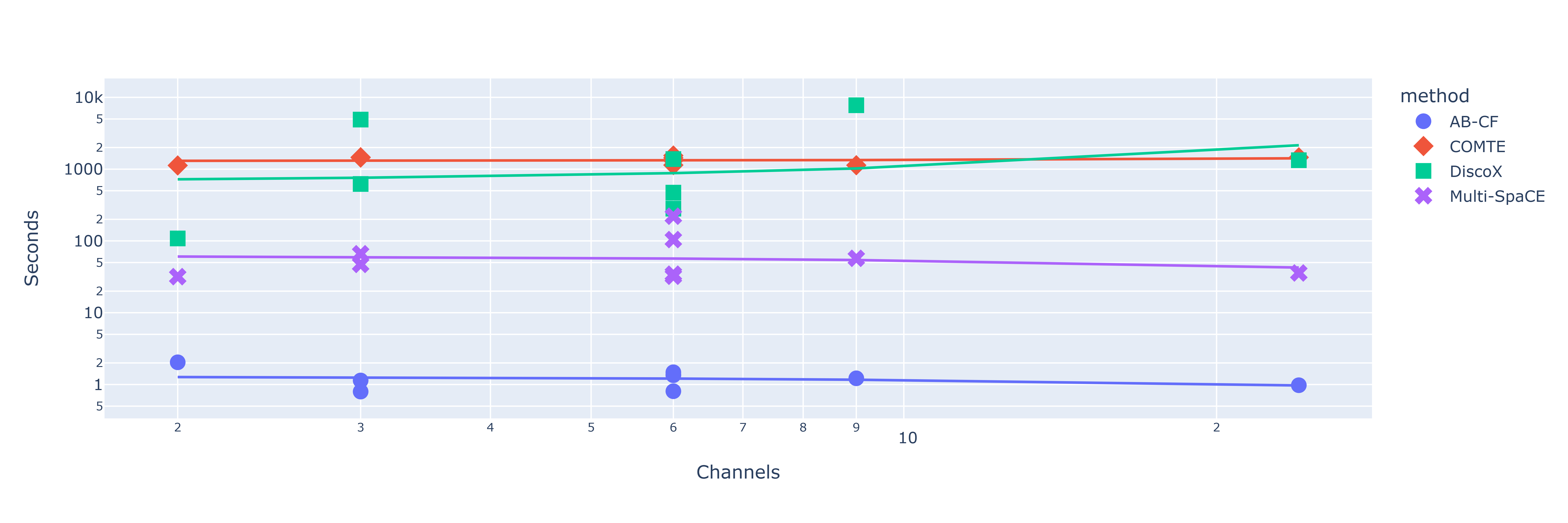}
    \end{subfigure}
    \caption{Execution times depending on the length and the number of channels in multivariate datasets.}
    \label{fig:multivariate_times}
\end{figure}

\begin{figure}[h]
    \centering
    
    \includegraphics[width=0.9\textwidth]{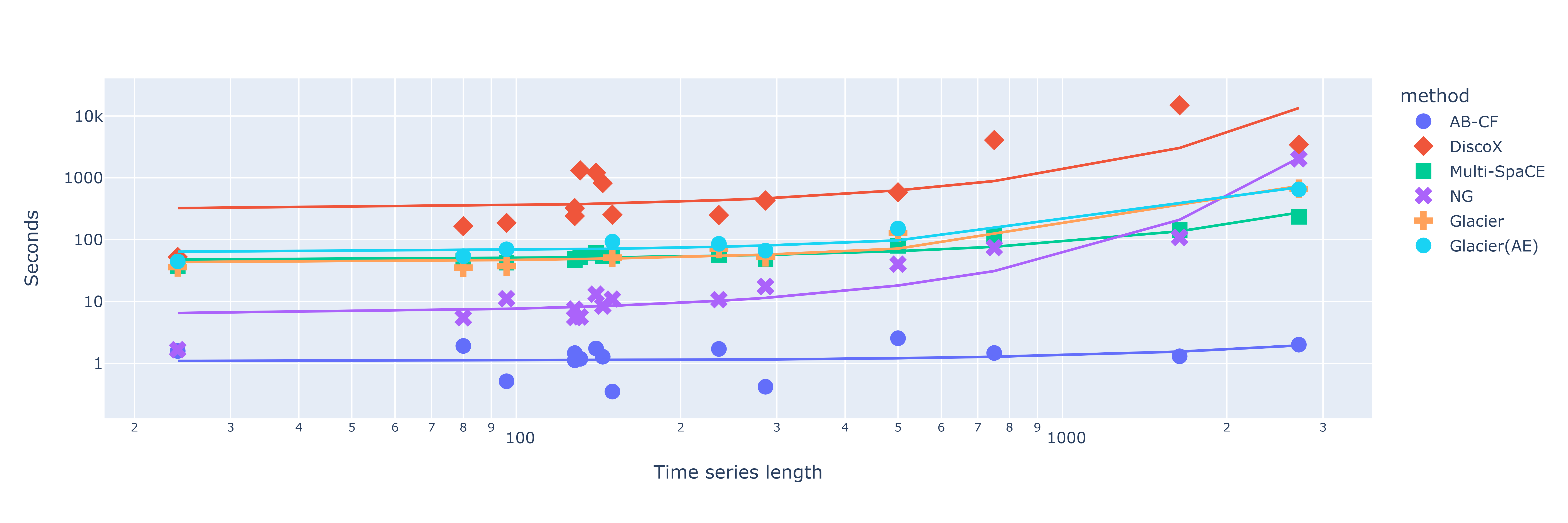}

    \caption{Execution times depending on the length in univariate datasets.}
    \label{fig:univariate_times}
\end{figure}

We analyze the execution times of all the methods considered in the experiments. We rely on parallelization by dividing the 100 instances to be explained for each dataset into 20 chunks of 5 instances. Counterfactuals are then generated using 10 parallel processes for each method. Execution times are evaluated with respect to both the length of the time series and the number of input channels. To examine trends, we fit an ordinary least squares regression model, observing the relationship between input characteristics and execution times. Given the variability in dataset input dimensions, we use logarithmic scales for both the x- and y-axes. Results are presented in Figure~\ref{fig:multivariate_times} and Figure~\ref{fig:univariate_times}.

Multi-SpaCE generally falls in the middle of the spectrum, being faster than DiscoX, COMTE, and Glacier, but slower than NG and AB-CF. The results reveal that, in general, execution times are more sensitive to increases in time series length than to the number of input channels.

Due to multiprocessing, we observed typical overheads, including bottlenecks caused by GPU sharing across processes, which resulted in imperfect parallel executions. Consequently, the reported execution times per method may not exactly reflect those expected in a standard environment without parallelization. To evaluate this effect, we repeated the experiments on multivariate datasets without parallelization. The results of this experiment are reported in Table~\ref{tab:times_parallel_sequential}.
\begin{center}
\setlength{\tabcolsep}{3pt}
\setlength\extrarowheight{-3pt}
\begin{table}[H]
\centering
\resizebox{0.55\textwidth}{!}{
    \begin{tabular}{l|cc}
    \hline
    Dataset & Multi-SpaCE(parallel) & Multi-SpaCE \\
    \hline
    AWR & 57.27 & 26.67 \\
    BasicMotions & 34.45 & 21.86 \\
    Cricket & 220.96 & 94.67 \\
    Epilepsy & 47.11 & 21.98 \\
    NATOPS & 35.81 & 18.79 \\
    PEMS-SF & 408.85 & 178.45 \\
    PenDigits & 31.82 & 16.93 \\
    RacketSports & 32.23 & 17.67 \\
    SR-SCP1 & 104.44 & 42.41 \\
    UWave & 66.32 & 25.61 \\
    \hline
    \end{tabular}
}
\caption{Average execution time (in seconds) when using normal and parallel experimentation.}
\label{tab:times_parallel_sequential}
\end{table}
\end{center}

\end{appendices}

\end{document}